\documentclass[lettersize,journal]{IEEEtran}
\usepackage{amsmath,amsfonts}
\usepackage{xcolor}
\usepackage{algorithmic}
\usepackage{algorithm}
\usepackage{array}
\usepackage[caption=false]{subfig}
\usepackage{textcomp}
\usepackage{stfloats}
\usepackage{multirow}
\usepackage{url}
\usepackage{ulem}
\usepackage{verbatim}
\usepackage{graphicx}
\usepackage{cite}
\usepackage{threeparttable}
\usepackage{ulem} 
\begin{document}

\title{TON-VIO: Online Time Offset Modeling Networks for Robust Temporal Alignment in High Dynamic Motion VIO}
\author{Chaoran Xiong,~\IEEEmembership{Student~Member,~IEEE}, Guoqing Liu, Qi Wu,~\IEEEmembership{Student~Member,~IEEE}, Songpengcheng Xia,~\IEEEmembership{Graduate~Student~Member,~IEEE}, Tong Hua, Kehui Ma, Zhen Sun, \\ Yan Xiang and Ling Pei$^{\ast}$,~\IEEEmembership{Senior~Member,~IEEE} 

\thanks{This work was supported in part by the National Natural Science Foundation of China (NSFC) under Grant No.62273229, No.61873163 separately
and in part by smart City beidou spatial-temporal digial base construction
and application industrialization(HCXBCY-2023-020).}

\thanks{$^{\ast}$Corresponding author: Ling Pei.}
\thanks{The authors are with the Shanghai Jiaotong University, Shanghai 200240,
China (e-mail: sjtu4742986; guoqing\_liu; robotics\_qi; songpengchengxia; ht1234; khma0929; zhensun; yan.xiang; ling.pei@sjtu.edu.cn).}

}

\markboth{Journal of \LaTeX\ Class Files}%
{Shell \MakeLowercase{\textit{et al.}}: A Sample Article Using IEEEtran.cls for IEEE Journals}


\maketitle

\begin{abstract}
Temporal misalignment (time offset) between sensors is common in low cost visual-inertial odometry (VIO) systems. Such temporal misalignment introduces inconsistent constraints for state estimation, leading to a significant positioning drift especially in high dynamic motion scenarios. In this article, we focus on online temporal calibration to reduce the positioning drift caused by the time offset for high dynamic motion VIO. For the time offset observation model, most existing methods rely on accurate state estimation or stable visual tracking. For the prediction model, current methods oversimplify the time offset as a constant value with white Gaussian noise. However, these ideal conditions are seldom satisfied in real high dynamic scenarios, resulting in the poor performance. In this paper, we introduce online time offset modeling networks (TON) to enhance real-time temporal calibration. TON improves the accuracy of time offset observation and prediction modeling. Specifically, for observation modeling, we propose feature velocity observation networks to enhance velocity computation for features in unstable visual tracking conditions. For prediction modeling, we present time offset prediction networks to learn its evolution pattern. To highlight the effectiveness of our method, we integrate the proposed TON into both optimization-based and filter-based VIO systems. Simulation and real-world experiments are conducted to demonstrate the enhanced performance of our approach. Additionally, to contribute to the VIO community, we will open-source the code of our method on: https://github.com/Franky-X/FVON-TPN.
\end{abstract}

\section{Introduction}
\IEEEPARstart{A}{ccurate} state estimation is a fundamental issue for autonomous vehicles. Various state estimation algorithms have been developed based on different sensor sets\cite{Multi-Modal, LVSM, LiTao3, Hua2023}. One of the most typical and low-cost state estimation systems is visual-inertial odometry(VIO), which utilizes visual observations to constrain the inertial measurement unit(IMU) measurements\cite{VINS-Mono,OpenVINS, ORB-SLAM, CEVIO, Qi2023}.
\par
In a low-cost VIO system, there often exists a time-varying temporal misalignment between visual and IMU measurements due to different clock sources and data transmission latency\cite{Rehder2016}. Even with a synchronized external trigger, there is no guarantee that the sample moments of camera and IMU are strictly aligned because of their different sampling process. For instance, in most raw vehicle datasets\cite{KAIST}, \cite{ZOD}, there is usually 10-50 ms time-varying time offset between camera and IMU measurements.
\par
The temporal misalignment usually leads to inconsistent visual constraints on IMU propagation. Due to the inconsistent constraints, the estimator converges to the fake optimal state, so that the state estimation drift is caused. This drift becomes particularly significant in high dynamic scenarios, as the high dynamic motion enlarges the inconsistency of visual constraints caused by the temporal misalignment. As a result, to reduce the estimation drift attributed to the time-varying and unknown time offset, online temporal calibration is an essential issue for VIO.
\par
Existing online temporal calibration schemes for VIO can be divided into state-relevant methods and state-irrelevant methods. The state-relevant methods incorporate time offset into IMU propagation to align with visual observations\cite{OpenVINS}, \cite{Li2014}. However, these methods rely on precise state estimation, particularly linear and angular velocity. On the other hand, the state-irrelevant methods modify the visual projection model by shifting features with their velocity on the image plane to align with IMU measurements\cite{qin2018, Liu2020}. Whereas, the computation of each feature's velocity depends on continuous and stable visual tracking condition. 
\par
For high dynamic motion VIO, accurate state estimation especially for linear and angular velocity is difficult to maintain, so that the state-relevant online temporal calibration methods are often unstable. By contrast, the state-irrelevant methods are more stable, but the poor visual tracking condition challenges the valid computation for the velocity of features. Consequently, the convergence of these methods is usually slow, causing accumulated positioning errors. Furthermore, few methods consider the temporal attributes of the time offset in online estimation. Hence, existing methods often perform poorly in time offset estimation in  high dynamic motion scenarios. To date, there has been limited research on robust temporal calibration methods specifically for high dynamic vehicle applications.
\par
In this paper, we focus on the robust estimation of the time offset between camera and IMU for high dynamic motion VIO. 
We propose novel online time offset modeling networks (TON) to enhance temporal calibration so that the drift of state estimation caused by temporal misalignment is effectively reduced. The proposed TON can be integrated into different optimization-based and filter-based VIO systems to improve the time offset estimation and overall positioning performance.
Our contributions are:
\begin{itemize}
  \item Novel online weakly-supervised learning networks TON for the time offset modeling that enhances temporal calibration for time-varying temporal misalignment for high dynamic motion VIO; 
  \item Feature velocity observation networks (FVON) integrated in the visual factor to effectively infer the velocity of poorly tracked features for the time offset observation modeling;
  \item Time offset prediction networks (TPN) learning the time offset evolution over time for its prediction modeling;
  \item Simulation and real-world experiments on various vehicles in different high dynamic motion challenging scenarios to demonstrate the effectiveness of our method compared with the state-of-the-art approaches.
\end{itemize}

The rest of this paper is organized as follows: Section II reviews current temporal alignment and calibration methods based on classic methods and deep learning networks. Section III formulates the key problems of online temporal calibration in real-case high dynamic motion VIO. Section IV presents our proposed online weakly-supervised learning TON to enhance temporal calibration.  Section V describes and discusses the experimental sets and results. Finally, Section VI concludes our proposed method with a summary.

\section{Related Work}
In this section, we provide an overview of the evolving field of time offset estimation. Section II-A discusses traditional techniques for reducing and estimating the time offset between sensors. On the other hand, Section II-B explores methodologies that leverage deep learning networks for temporal calibration.
\subsection{Classic Methods for Temporal Alignment and Calibration}
\par
Traditionally, temporal alignment is achieved by hardware or software synchronization. Hardware synchronization often employs an extra synchronization board to simultaneously trigger the sensors. \cite{Tschopp2020} developed an open-sourced hardware named Versa-VIS to trigger cameras with exposure time compensation and IMU by a unified clock source. For evaluation, a hardware-based synchronization accuracy evaluation tool was proposed in \cite{Yuan2021}, which amplified the tiny time offset into larger measurements. However, even if the hardware trigger delay is reduced to the microsecond level by a microchip with precise timing, there is no guarantee that all the sensors sample at the same time due to the unique timing feature affected by data acquisition process and transmission latency of each device.
\par
Alternatively, researchers propose software calibration solutions. Some focus on offline calibration, incorporating the time offset into the calibration parameters. Paul et al. proposed the well-known open-source tool Kalibr\cite{Rehder2016}, utilizing B-Spines to fit the continuous trajectory of IMU to align with the visual chessboard observations. Similarly, \cite{Zhu2022, Li2023, Han2021} also aligned the continuous trajectory fit by B-Spline of IMU to observations of other different modalities such as LiDAR. There were also researches utilizing B-Spine to fit the features' movement to align with the IMU measurements, including \cite{Yoon2022, Nowicki2020}. Moreover, loosely-coupled unified time offset estimation methods using trajectory correlation were proposed and verified by \cite{Qiu2021}. Nevertheless, the offline methods above could only provide an initial, but not accurate, value for the time offset in a real VIO system because the time offset is a time-varying value evolving over time.
\par
Therefore, some researchers focus on online temporal calibration methods. Li et al. were the first to consider online temporal calibration within the extended Kalman filter framework of VIO\cite{Li2014}. They took time offset into the IMU pre-integration to estimate it online as a parameter and analyzed the observability of the system. \cite{Yang2022, Huai2022, Yang2023, Li2020} applied the method into different form of filters including adaptive Kalman filter, multi-state constraint Kalman filter, and unscented Kalman filter. Apart from filter-based VIO, the idea of taking time offset into the IMU pre-integration was also introduced to the optimization-based VIO framework by \cite{Fu2022, Wang2022_1, Yan2023}. However, these methods could only achieve high-accuracy time offset estimation based on precise state estimation, especially linear and angular velocity. In high dynamic motion scenarios, the accurate state estimation is challenging to maintain, so that methods relying on precise state estimation perform poorly. Thus, Qin et al. first proposed a state-irrelevant method for time offset estimation in VIO\cite{qin2018}. They observed the time offset on the image plane by the velocity of features. The movement of features on the image plane was taken to align with IMU measurements. However, the feature velocity computation only worked efficiently under ideal visual observation conditions due to the constant speed model assumption, but performed poorly in high dynamic vehicle applications.
\par
Furthermore, few studies on time offset estimation have analyzed the evolution pattern of the time offset over time. This pattern is a crucial aspect of the time offset and should be considered in the estimation constraints.

\subsection{Learning Methods for Temporal Calibration}
Few studies facilitate deep neural networks for time offset calibration. To date, self-supervised learning methods have been used primarily in LiDAR-camera calibration. As these sensors share a common view, features can be aligned in the same modal domain. Yuan et al. proposed a self-supervised end-to-end LiDAR-camera alignment model \cite{Yuan2022}, utilizing stereo depth estimation as labels for supervising the training process. Additionally, there are end-to-end deep learning methods in VIO \cite{VINET}. The temporal misalignment is reduced by certain deep neural networks. However, these end-to-end methods often lack generality, and the exact value of the time offset cannot be extracted. 

\begin{figure*}
  \centering
  \includegraphics[width=0.8 \textwidth]{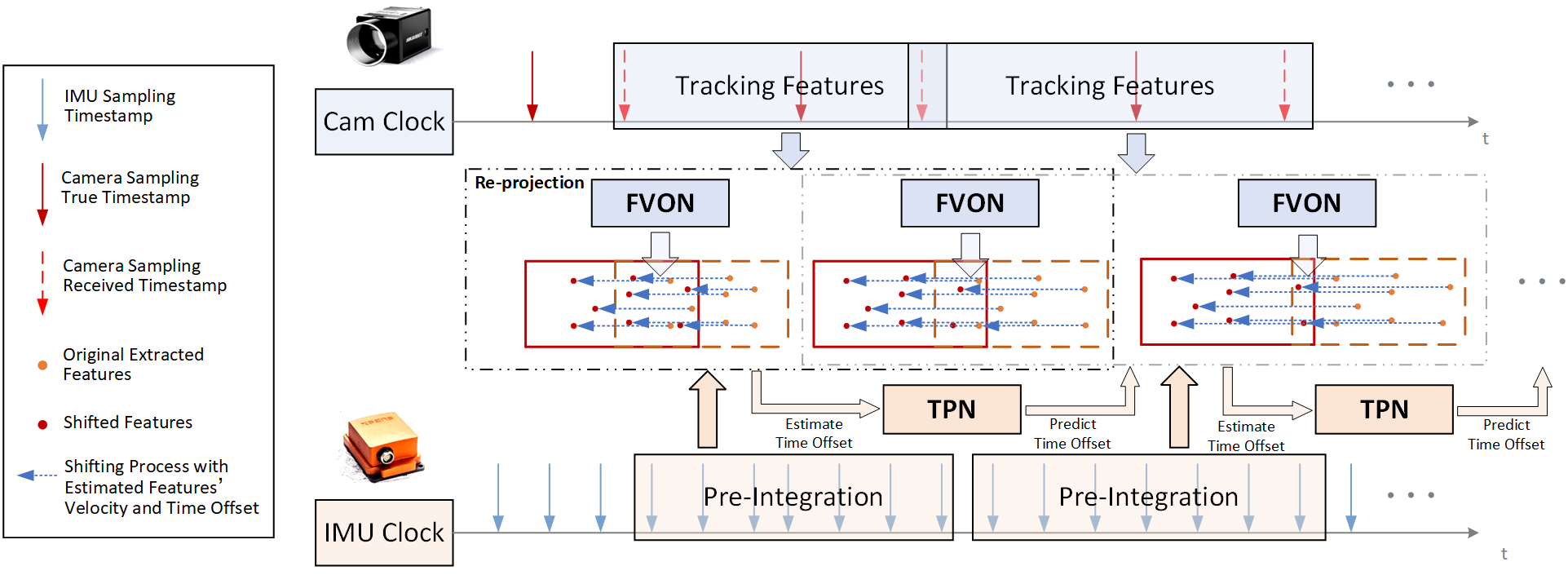}
  \caption{System overview of our proposed online weakly-supervised learning TON composed of FVON and TPN enhancing online temporal calibration. The system inputs are raw camera and IMU data with time-varying offset between them. First, the pre-integration for IMU data is implemented according to the received timestamps of camera and IMU. At the same time, the visual front-end extracts and tracks features between consecutive frames. Second, the IMU pre-integration results and front-end tracked features are incorporated into a solver to perform re-projection. During the solving process, the proposed weakly-supervised FVON and TPN enhance the observation and prediction modeling of time offset estimation.}
  \label{fig:System Framework}
\end{figure*}

\section{Problem Formulation}
\subsection{Observation Model for Time Offset}
In high dynamic motion scenarios, accurate state estimation is challenging to maintain for the state-relevant methods. Hence, we focus on the more stable state-irrelevant method. The state-irrelevant time offset observation is incorporated into the projection model by shifting features at their velocity on the image plane\cite{qin2018}. The incorporation is depicted as follows:
\begin{equation}
    \mathbf{z}_l^{k}(t_d)=\mathbf{z}_l^{k}-\mathbf{V}_l^kt_d,
\end{equation}
where $t_d$ is the estimated time offset. $ \mathbf{z}_l^{k}(t_d)$ and $\mathbf{z}_l^{k}$ are the shifted and original position of the feature, respectively. $\mathbf{V}_l^k$ is the velocity of feature $l$ in frame $k$.
\par
\begin{figure}
  \centering
  \includegraphics[width=0.5 \textwidth]{"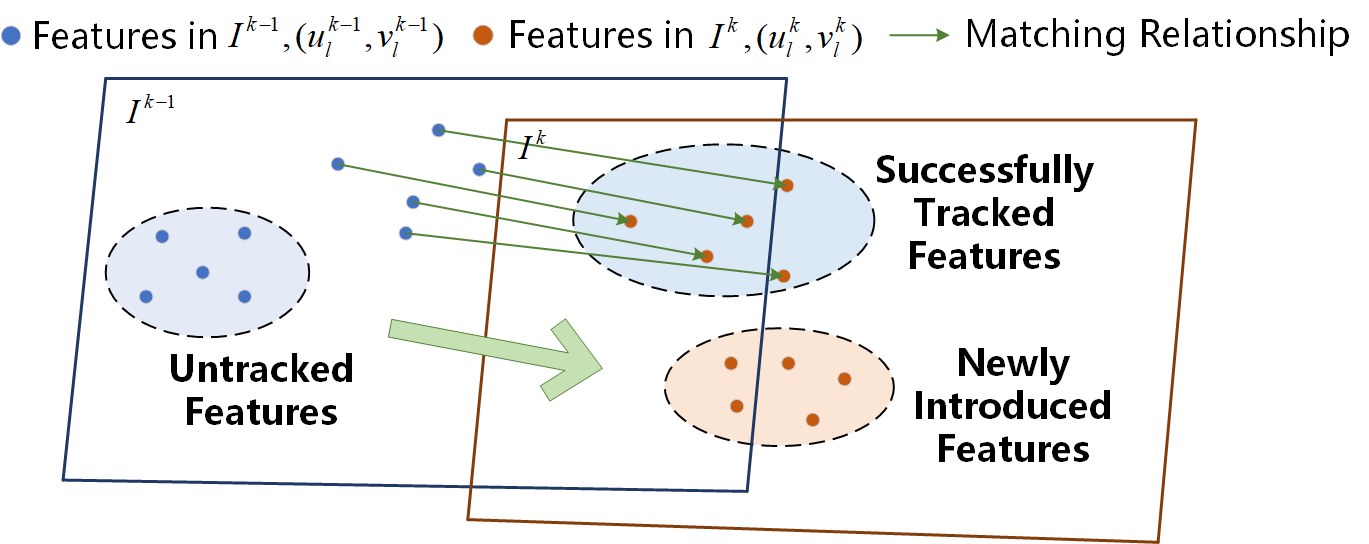"}
  \caption{The observed features in two consecutive frames. The blue points represent the features observed in $I^{k-1}$ frame, while the orange points represent the features observed in $I^{k}$ frame. The matching relationship is illustrated as arrows. The traditional features' velocity computation method is only capable of estimating the velocity of features which have previous observation, while failing to calculate the newly introduced features in $I^{k}$ frame due to lack of valid tracking in previous frame.}
  \label{fig:FVO}
\end{figure}

In the projection model above, the velocity of feature $l$ in frame $k$ is the key factor for the observation model of the time offset. As depicted in Fig. \ref{fig:FVO}, the computation of the feature velocity is traditionally based on the constant speed model\cite{qin2018} shown as follows:
\begin{equation}
\mathbf{V}_l^k=\left(\left[\begin{array}{c}
u_l^{k} \\
v_l^{k}
\end{array}\right]-\left[\begin{array}{c}
u_l^{k-1} \\
v_l^{k-1}
\end{array}\right]\right) /\left(t_{k}-t_{k-1}\right),
\end{equation} \\
where $\mathbf{V}_l^k$ is the velocity of feature $l$ in frame $k$. $\left[u_l^k, v_l^k\right]^T$ and $\left[u_l^{k-1}, v_l^{k-1}\right]^T$ are the observations of feature $l$ on the image plane $I^{k}$ and previous $I^{k-1}$, respectively. However, the absence of prior tracking for features significantly challenges the computation of features' velocity, especially in high dynamic motion scenarios. Due to insufficient valid computations of features' velocity, the observation model of the time offset is inaccurate, leading to slow convergence and eventually significant drift in positioning.
\par
The problem is to modify the computation of features' velocity in unstable visual condition, where continuous feature tracking is challenging. Our aim is to use weakly-supervised learning techniques for improving the computation of features' velocity in the projection model, so that more accurate observation model for time offset estimation can be achieved. 

\subsection{Prediction Model for Time Offset}
In a typical low-cost camera and IMU hardware set, due to various clock sources, random data transmission latency, and different data sampling processes, a time-varying time offset between sensors often exists. The time offset usually evolves over time according to an unknown pattern. In previous works\cite{OpenVINS},\cite{Li2014},\cite{qin2018}, the prediction model of the time offset is assumed to be a constant value with white Gaussian noise to simplify the problem, which is inaccurate for real-case time-varying time offset estimation. In this paper, we define the evolution of the time offset between camera and IMU closer to the actual physical process as follows:
\begin{equation}
t_{cam}(k) = t_{imu}(k) + t_{d}(k),
\end{equation}
\begin{equation}
t_{d}(k) = t_d(k-1) + t_{bias} + n_{t_d},
\end{equation}
where $t_{cam}(k)$ and $t_{imu}(k)$ denote the actual sampling timestamp of camera and IMU in the frame $k$. $t_d(k)$ represents the time offset between $t_{cam}(k)$ and $t_{imu}(k)$. $t_d(k+1)$ and $t_d(k)$ are the time offset of $(k+1)$th and $k$th frame, respectively. The time offset suffers from a cumulative bias drift $t_{bias}$ and white Gaussian noise $n_{t_d}$ because of the sampling frequency difference of different sensors. As a result, predicting the evolution of the time offset based on a traditional model is challenging because of its uncertain parameters of different devices.
\par
The problem is to model the unknown and time-varying time offset between sensors online. Our goal is to utilize weakly-supervised learning techniques to learn the evolution pattern of time offset to improve the prediction model for online temporal calibration. 

\section{Methodology}
\par
In this section, we first provide an overview of our proposed online weakly-supervised learning TON that enhances the time offset model for temporal calibration. Next, we present the two key components of TON: FVON and TPN. Finally, we discuss the implementation of the proposed FVON and TPN within both optimization-based and filter-based VIO systems.

\subsection{System Overview}
The designed online weakly-supervised learning TON for temporal calibration is depicted in Fig. \ref{fig:System Framework}. The system inputs are raw camera and IMU data with time-varying offset between them. First, the pre-integration of IMU data is performed according to the received timestamps from the camera and IMU. Simultaneously, the visual front-end extracts and tracks features between consecutive frames. Next, the results of IMU pre-integration and the front-end tracked features are incorporated into a solver for re-projection. During the solving process, the proposed weakly-supervised FVON and TPN enhance the modeling of time offset observation and prediction. FVON is designed to predict the velocity of poorly tracked features, allowing for more precise feature shifting on the image plane for time offset observation modeling. TPN is designed to learn the evolution pattern of time offset over time for prediction modeling. Both FVON and TPN are trained online using weakly-supervised labels acquired from current and past visual trackings and states. In the next subsections, we provide detailed elaboration on the proposed FVON and TPN.
\subsection{FVON: Feature Velocity Observation Networks}
The state-irrelevant online temporal calibration method relies on the velocity of features for time offset observation modeling. In high dynamic motion scenarios, each frame has to introduce a substantial number of new features to acquire sufficient visual constraints. However, conventional computation of features' velocity, relying on previous observations, falls short in dealing with new features that lack prior observations. The lack of valid velocity constraints leads to inaccurate time offset observation modeling. Therefore, to enhance velocity constraints of features for time offset observation, we design FVON to predict the velocity of poorly tracked features. FVON consists of two modules: 1) inverse time series-FVON and 2) frame-to-frame-FVON. 
\subsubsection{Inverse Time Series (ITS)-FVON}

\begin{figure}
  \centering
  \includegraphics[width=8 cm]{"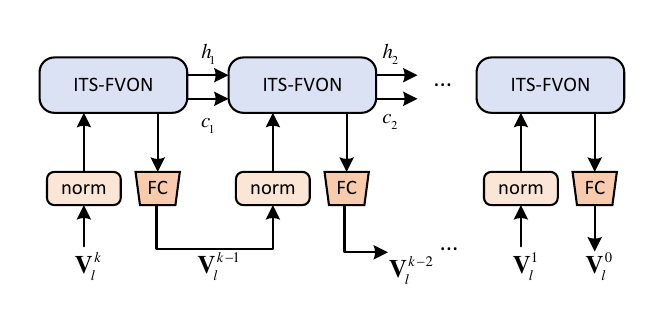"}
  \caption{ITS-FVON architecture. The features' velocity of future frames is normalized to $[0,1]$ and then input to the ITS-FVON. The hidden state is used for passing the temporal relationship of future, present and previous features' velocity.The output is the prediction the features' velocity of previous frame.}
  \label{fig:LSTM_V}
\end{figure}

For the velocity prediction of features lacking previous observation but tracked in the next few frames, we introduce a solution: inverse time series features' velocity observation weakly-supervised networks. Below, we present the problem definition:
\begin{equation}
\mathbf{V}_l^k=F_{w}(k).
\end{equation}
In each window, we assume that a feature's velocity $\mathbf{V}_l^k$ in frame $k$ can be expressed as a non-linear function $F_{w}(k)$. To compute the velocity of features lacking previous observations, we rely on the velocity of other well-matched features within the time series for prediction. Typically, when a feature is consistently tracked in recent frames, we utilize the latter features' velocity to estimate the initial one, which lacks previous observation. As a result, approximating the non-linear temporal evolution function of features' velocity becomes a modeling problem. To this end, we propose a neural network, ITS-FVON, defined as follows:
\begin{equation}
\text{ITS-FVON}: \mathbf{V}_l^{k+1} \mapsto\mathbf{V}_l^k,
\label{ITS-FVON}
\end{equation}
where the input is the features' velocity in the frame $(k+1)$, denoted as $\mathbf{V}_l^{k+1}$, and the output is the estimated feature's velocity in the frame $k$, denoted as $\mathbf{V}_l^k$.
\par
The input to the ITS-FVON is the feature's velocity in frame $(k+1)$ and normalized in the solving window to the range $[0,1]$. Then the normalized $\mathbf{V}_l^{k+1}$ is input to the ITS-FVON with hidden states $\mathbf{h}_i$ and $\mathbf{c}_i$ passing through the future, present, and previous frames for the estimation of features' velocity.
The loss function of ITS-FVON is defined by the mean square error (MSE) as follows:
\begin{equation}
\mathcal{L}(\mathbf{V}_l^k, \hat{\mathbf{V}}_l^k)=\frac{1}{n} \sum_{k=1}^n\left\|\mathbf{V}_l^k-\hat{\mathbf{V}}_l^k\right\|^2,
\label{Loss}
\end{equation}
where $\mathcal{L}(\mathbf{V}_l^k, \hat{\mathbf{V}}_l^k)$ represents the loss function. $\mathbf{V}_l^k$ and $\hat{\mathbf{V}}_l^k$ are the labeled feature's velocity and the feature's velocity predicted by ITS-FVON, respectively. $n$ is the number of training labels.
\par
\begin{figure}
  \centering
  \includegraphics[width= 0.4 \textwidth]{"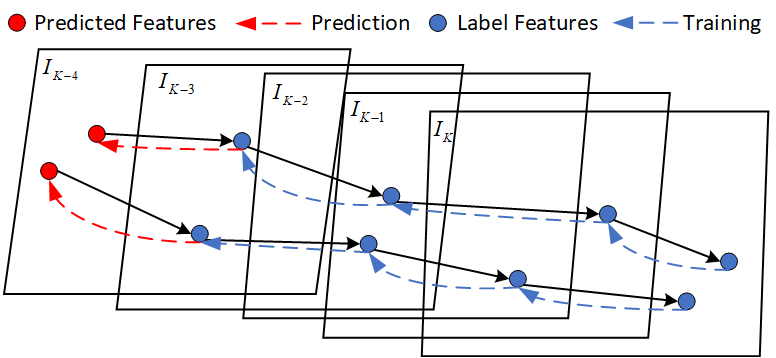"}
  \caption{ITS-FVON weakly-supervised training strategy. The blue points represent features which have previous observations. In other words, its velocity can be calculated with traditional method. We take these features as labels to train the network. The red points represent features of which the velocity cannot be calculated with traditional method due to the lack of previous valid observation. The trained network is used for its prediction.}
  \label{fig:LSTM_Label}
\end{figure}

The network is trained using (\ref{Loss}) to predict features' velocity. As illustrated in Fig. \ref{fig:LSTM_Label}, the labels are obtained from well-matched features whose velocity can be calculated using traditional methods within the time sequence of the window. Once trained, the ITS-FVON is applied to predict velocity of features lacking consecutive observations.

\par

\subsubsection{Frame-to-Frame (F2F)-FVON}
In certain scenarios, features lacking previous observations cannot be tracked over extended periods. In such instances, ITS-FVON may struggle to provide precise estimates of the features' velocity. To tackle this issue, we propose a solution: frame-to-frame weakly-supervised neural networks for features' velocity observation. Below, we present the problem definition:
\begin{equation}
\mathbf{V}_l^k=F_{\theta}(\mathbf{P}_l^k).
\end{equation}

We assume that the velocity of feature $l$ in frame $k$ can be represented as a non-linear function $F_{\theta}(\mathbf{P}_k^l)$, where $\mathbf{P}_k^l$ is the three-dimensional (3D) position of feature $l$ in frame $k$.
Between two consecutive frames, the velocity of each feature $l$ can be computed through the kinematic motion as follows:
\begin{equation}
\mathbf{V}_l^k= \frac{\text{d}(\pi(\mathbf{R}(t)\mathbf{P}_l^k+\mathbf{T}(t))}{\text{d}t},  
\end{equation}
 where $\mathbf{R}(t) $ is the rotation and $\mathbf{T}(t)$ is the translation of the camera over time. $\pi(\cdot)$ is the projection model of the camera. If we have enough valid features' velocity observations $\mathbf{V}_l^k$, the motion $\mathbf{R}(t),\mathbf{T}(t)$ can be solved by the above differentiate formula. Our proposed F2F-FVON is designed to estimate the motion of features between these two frames. During the interval between two consecutive frames, the features' velocity maintains a consistent pattern of motion. For computing velocity in cases where features are tracked for a short period, we introduce F2F-FVON specifically designed to capture the underlying motion patterns between two consecutive frames. The proposed neural network F2F-FVON is defined as follows:
\begin{equation}
\text{F2F-FVON}:\mathbf{P}_l^k \mapsto\mathbf{V}_l^k,
\end{equation}
where the input is the 3D position of feature $l$ in frame $k$, denoted as $\mathbf{P}_l^k$, and the output is the estimated velocity of feature $l$ in frame $k$.

The network is trained using equation (\ref{Loss}) for the velocity of features. Labels are provided by well-matched features whose velocities can be accurately calculated using traditional methods between two consecutive observations. The trained F2F-FVON is used for features lacking consecutive observations.

\begin{figure}
  \centering
  \includegraphics[width= 0.5 \textwidth]{"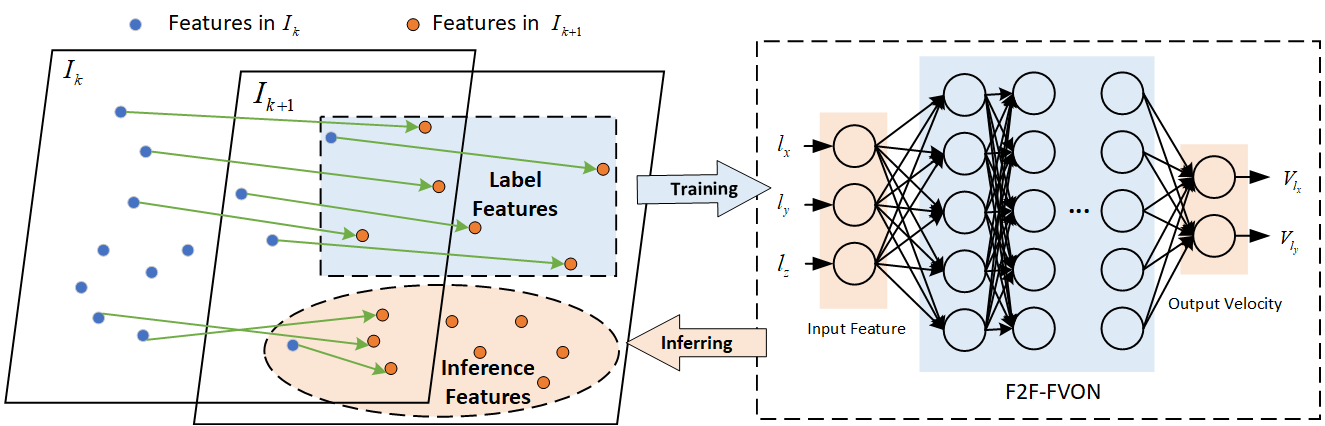"}
  \caption{F2F-FVON weakly-supervised training strategy. The blue points represent features with previous observation. Their velocity can be calculated with traditional method. These features are taken as labels to train the network. The orange points represent features of which the velocity cannot be calculated with traditional method due to the lack of valid observation. The trained F2F-FVON is used for their prediction.}
  \label{fig:MLP_Tra}
\end{figure}

\subsection{TPN: Time offset Prediction Networks}
In traditional time offset estimation within a VIO system, the time offset is assumed as a constant value and estimated independently in each window's estimation. However, in reality, the time offset evolves over time following a pattern with unknown parameters \cite{Tschopp2020}. Thus, the temporally independent estimation and constant value assumption of the conventional method conflict with the actual physical process of the time offset's evolution over time, leading to fragile estimation especially in high dynamic motion scenarios. Our goal is to incorporate the timing pattern of the time offset to enhance the robustness of its estimation. 
To this end, we propose TPN, weakly-supervised time offset prediction neural networks, defined as follows:
\begin{equation}
\text{TPN}:t_d(K) \mapsto t_d(K+1),
\end{equation}
where the input is the estimated time offset $t_d$ in window $K$, denoted as $t_d(K)$, and the output is the predicted time offset $t_d$ in window $(K+1)$, denoted as $t_d(K+1)$.

\begin{figure}
  \centering
  \includegraphics[width=8 cm]{"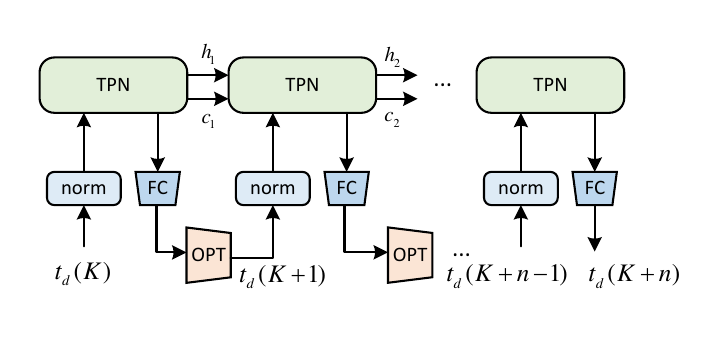"}
  \caption{TPN architecture. The time offset of the previous optimization window is normalized to $[0,1]$. The normalized $t_d(K)$ is passed to the TPN then. The hidden state is used to pass temporal relationship information of time offset between the batch windows to the previous, present and future time offset estimation. The output is time offset prediction of the next batch window.}
  \label{fig:2}
\end{figure}

The input to TPN is the time offset estimated in window $K$, normalized to the range $[0,1]$ using data from previous windows. The normalized $t_d(K)$ is then passed to TPN with states $\mathbf{h}_i$ and $\mathbf{c}_i$, which conveys temporal information of time offset to the previous, present and future estimation.
The loss function of TPN is defined by MSE as follows:
\begin{equation}
\mathcal{L}(t_d(K), \hat{t}_d(K))=\frac{1}{n} \sum_{k=1}^n\left\|t_d(K)-\hat{t}_d(K)\right\|^2,
\label{Loss_td}
\end{equation}
where $\mathcal{L}(t_d(K), \hat{t}_d(K))$ represents the loss function. $t_d(K)$ and $\hat{t}_d(K)$ are the label time offset and predicted time offset by TPN, respectively. $n$ is the number of training labels.
\par
The network is trained using equation (\ref{Loss_td}) for time offset estimation. Labels are provided by the previously estimated time offset, derived from the solver. The trained network is then used for time offset estimation in the subsequent window.

\subsection{Implementation of FVON and TPN in VIO}
\subsubsection{Networks Configuration}
Table I outlines the configuration of our proposed networks. We utilize long short-term memory networks (LSTM) with a single layer and 2 hidden states for the ITS-FVON, 4-layer multi-layer perceptron (MLP) with 5 hidden states for F2F-FVON, and a single-layer LSTM with 2 hidden states for TPN.
\begin{table}[H] \footnotesize
\renewcommand\arraystretch{1.2}
    \centering
    \caption{Configuration of Proposed Networks}
    \begin{tabular}{|c|c|c|c|}
    \hline
     \textbf{Proposed Networks}    &   \textbf{Layer}  &  \textbf{LR} & \textbf{Epoch}\\
     \hline
     ITS-FVON    &    LSTM(2,2,2)  &   0.0001 & 1000 or loss $<$ 1e-5 \\
     \hline
     F2F-FVON    &    MLP(3,5,5,2) & 0.01 & 1500 \\
     \hline
     TPN         &    LSTM(1,2,1)  &  0.0001 & 1000 or loss $<$ 1e-8 \\
     \hline
    \end{tabular}
    \label{tab:my_label}
\end{table}
The proposed FVON and TPN are designed to run efficiently on low-grade GPUs in laptop configurations.
\subsubsection{System Integration}
We integrate our proposed FVON and TPN into the VIO systems in four stages to enhance time offset estimation. In stage 1, the dataset for FVON training labels is loaded during the feature tracking process of VIO with well-matched features' velocity computed by the traditional constant speed model. In stage 2, either ITS-FVON or F2F-FVON is selected based on the tracking periods of features to predict the velocity of poorly tracked features. Next, the chosen module is trained online with the appropriate labels and subsequently used to predict the feature's velocity. The predicted feature's velocity is incorporated into the modified re-projection model to strengthen the time offset constraints, leading to faster convergence. In stage 3, the dataset for TPN training labels is loaded and updated with the previous time offset estimation  obtained from the solver. In stage 4, TPN is trained online to predict the time offset for the next estimation. 
\par
For the optimization-based VIO system, the time offset prediction from TPN is incorporated into the factor graph of the state estimator, where it is modeled as a unique factor that accounts for the time offset error state.
The residual for each batch window is defined as follows: 
\begin{equation}
\begin{aligned}
& \min _{\mathcal{X}}\{{\left\|\mathbf{e}_p-\mathbf{H}_p \mathcal{X}\right\|^2}+{\sum_{k \in \mathcal{B}}\left\|\mathbf{e}_{\mathcal{B}}\left(\mathbf{z}_{k+1}^k, \mathcal{X}\right)\right\|_{\mathbf{P}_{k+1}^k}^2} \\
& +\underbrace{\sum_{(l, j) \in \mathcal{C}}\left\|\mathbf{e}_{\mathcal{C}}\left(\mathbf{z}_l^j, \mathcal{X}\right)\right\|_{\mathbf{P}_l^j}^2}_{\text {with proposed \textbf{FVON} in the visual factor}}\ \\
& +\underbrace{\left\|t_{d}-t_{d}^{\text{TPN}}\right\|_{{P}_{t_d}}^2}_{\text {proposed \textbf{TPN} constraint }}\} .
\end{aligned}
\end{equation}

The residuals of an optimization window include the original IMU propagation error and prior information in the VIO framework, denoted as $\mathbf{e}_{\mathcal{B}}\left(\mathbf{z}_{k+1}^k, \mathcal{X}\right)$ and $\mathbf{e}_p-\mathbf{H}_p \mathcal{X}$, respectively. The residuals also include the modified visual projection error with our proposed FVON, which enhances time offset constraints for its observation model, and our uniquely designed TPN error for its predition model, denoted as $\mathbf{e}_{\mathcal{C}}\left(\mathbf{z}_l^j, \mathcal{X}\right)$ and $t_{d}-t_{d}^{\text{TPN}}$, respectively. The covariance of each factor is adjusted to different values for the visual residuals in each experiment.
\par
For the filter-based VIO system, the time offset prediction from TPN is incorporated into the state propagation process. The jacobian of the time offset is given as follows:
\begin{equation}
    J_{t_d} = (t_d^\text{TPN}(K) - t_d(K-1)) / d_t,
\end{equation}
where $J_{t_d}$ is the jacobian of $t_d$ prediction model. $t_d^\text{TPN}(K)$ is the predicted time offset of TPN. $t_d(K-1)$ is the previous time offset estimation. $d_t$ is the interval of two estimations.
\par
\begin{figure}
  \centering
  \includegraphics[width= 0.5 \textwidth]{"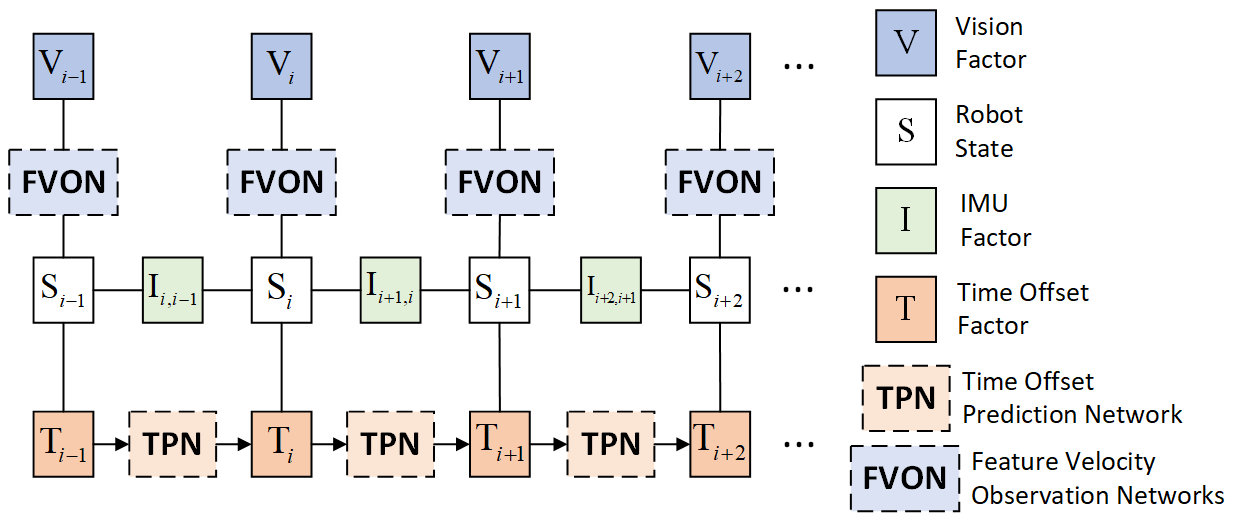"}
  \caption{Factor graph with proposed time offset prediction factor. The proposed FVON is integrated into the visual factor and weakly supervised by the present and future visual observations in a solving window. The proposed TPN predictions
are modeled as factors on the time offset error state and weakly supervised by history time offset estimation.}
  \label{fig:Factor}
\end{figure}

 The factor graph of our proposed TON is depicted in Fig. \ref{fig:Factor}, and the workflow of our FVON and TPN integration in a general VIO system is presented in Algorithm 1.

\begin{algorithm}
\caption{Proposed Algorithm.}
\begin{algorithmic}
\STATE \textbf{Stage 1: FVON dataset construction}
  \FOR{feature $l$ in frame $k$ of the current window}
    \IF{feature $l$ is well-matched}
      \STATE $\mathbf{V}_{l}^k$ = $\left(\left[\begin{array}{c}
u_l^{k} \\
v_l^{k}
\end{array}\right]-\left[\begin{array}{c}
u_l^{k-1} \\
v_l^{k-1}
\end{array}\right]\right) /\left(t_{k}-t_{k-1}\right);$
      \STATE Add $\mathbf{V}_{l}^k$ to labels;
    \ENDIF
  \ENDFOR
\STATE \textbf{Stage 2: FVON prediction}
 \FOR{Feature $l$ in frame $k$ of the current window}
    \IF{feature $l$ does not have previous observation}
        \IF{feature $l$ is tracked over 3 consecutive frames}
            \STATE Take velocity labels of feature $l$ in the next tracked frames to train ITS-FVON;
            \STATE $\mathbf{V}_{l}^k$ = ITS-FVON($\mathbf{V}_l^{k+1}$);
        \ELSE
            \STATE Take velocity labels of other features in frame $k$ to train F2F-FVON;
            \STATE $\mathbf{V}_{l}^k$ = F2F-FVON($\mathbf{P}_l^k$);
        \ENDIF
    \ENDIF
    \STATE Add $\mathbf{V}_{l}^k$ to the modified projection model for time offset observation;
 \ENDFOR
 \STATE \textbf{Stage 3: TPN dataset construction}
 \IF{the number of labels in the TPN dataset $>$ 30}
     \STATE erase the earliest label in the TPN dataset;
 \ENDIF
 \STATE Add the estimated time offset $t_d(K-1)$ of the previous window to labels; 
 \STATE \textbf{Stage 4: TPN prediction}
 \STATE Take time offset labels in TPN dataset to train TPN;
 \STATE $t_d^\text{TPN}(K) = \text{TPN}(t_d(K-1))$;
 \STATE Use $t_d^\text{TPN}(K)$ for an additional factor in optimization or propagation process in filter. 
\end{algorithmic}
\end{algorithm}

\section{Experiments}
In this section, we evaluate our proposed TON for online temporal calibration against other state-of-the-art methods. First, we introduce the experimental setup, design of evaluation metrics, and development of datasets for online temporal calibration validation. Then, we conduct various experiments to evaluate the effectiveness of our proposed method, with a focus on two main aspects: 1) the convergence and accuracy of time offset estimation, and 2) the overall positioning performance of VIO.

\subsection{Experimental Setup}
\subsubsection{Evaluation Metrics}
To evaluate the convergence of time offset estimation, timing error impact on state estimation, and overall positioning performance, we design the following three evaluation metrics.
\par
\textit{Convergent Iteration Times(CIT)}: We define CIT to assess the convergence speed of online temporal calibration methods as follows:
\begin{equation}
    \begin{aligned}
            \text{CIT} = \min_k \text{ such that } \left| \hat t_d(k) - t_{\text{target}} \right| \leq \epsilon_1 \\ 
            \& \left| \hat t_d(k) - \hat t_d(k-1) \right| \leq \epsilon_2,
    \end{aligned}
\end{equation}
where $\hat t_d(k)$ and $\hat t_d(k-1)$ are the estimated time offset in frame $k$ and $k-1$, respectively. $t_{\text{target}}$ is the target value for time offset convergence. $\epsilon_1$ is the error level. $\epsilon_2$ is the convergent stability level. The metric quantifies the minimum iterative times needed to achieve a certain error level.
\par
\textit{Timing Position Error (TPE)}: To quantify the impact of time offset on the positioning accuracy in a real VIO system, we introduce TPE. It evaluates how the time offset affects the position of visual observations, leading to state estimation errors. TPE is defined as follows:
\begin{equation}
\text{TPE}(k) = (\hat t_d(k) - t_{\text{gt}}(k)) \hat{\mathbf{v}}(k)
\end{equation}
where $\hat t_d(k)$ is the estimated time offset in frame $k$. $t_{\text{gt}}(k)$ is the ground truth time offset in frame $k$, and $\mathbf{v}(k)$ is the velocity of the state estimated in frame $k$. This metric quantifies the positioning error in observations due to the time offset estimation error, which is crucial for robust state estimation.
\par
\textit{Absolute Trajectory Error (ATE)}: ATE, including absolute position error and rotation error, is employed to assess the overall VIO performance influenced by online temporal calibration. This metric specifically evaluates the impact of these calibration methods on overall positioning accuracy.

\subsubsection{Development of Experimental Datasets}
To demonstrate the effectiveness of our proposed method, we conduct experiments in simulated and real-world sensor sets. The specifics of these experiments are detailed in Table I.
\par
\begin{figure}
  \centering
  \includegraphics[width= 0.5 \textwidth]{"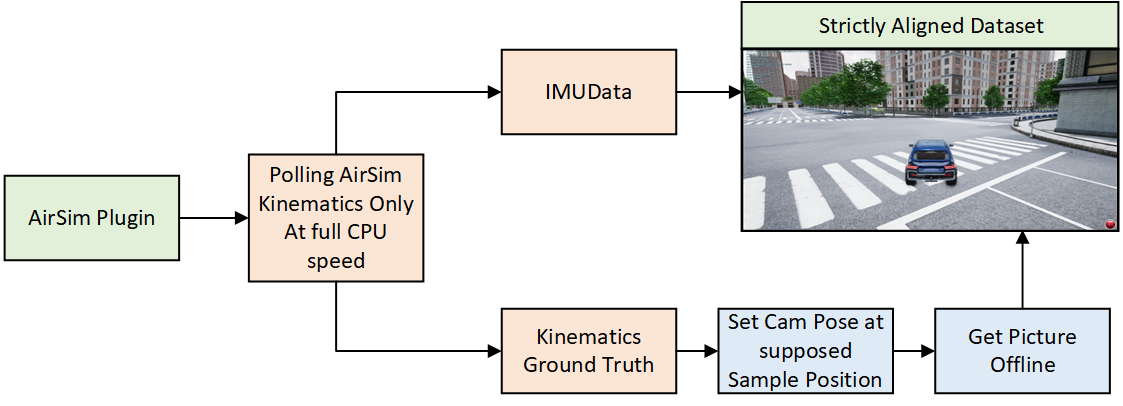"}
  \caption{AirSim fully synchronized data collection process. The first step is to acquire IMU data and ground-truth poses. The second step is to set the virtual camera at fixed ground-truth poses to capture images strictly aligned to IMU timestamps.}
  \label{fig:AirSim_Get}
\end{figure}

\textit{Simulation Dataset}: To accurately evaluate the time offset estimation of our methods, we develop a dataset with precisely known time offset ground truth. This dataset was generated using AirSim, a realistic simulator based on Unreal Engine 4, leveraging its virtual cameras and IMU. The data generation involves two steps, as shown in Fig. \ref{fig:AirSim_Get}: 1) Collecting IMU data and ground-truth poses, and 2) Offline capturing of fixed-pose images. This ensures a strict synchronization between visual and IMU data, essential for accurate time offset ground truth. We facilitate a car model in AirSim to create 2 high dynamic motion sequences for thorough evaluation.
\par
\textit{Public Open EuRoC Dataset}\cite{EuRoC}: The EuRoC dataset is popular in the field of visual-inertial navigation due to its ideal visual observations and high-quality IMU measurements. It features a micro aerial vehicle (MAV) operating in six-degree freedom indoor environments. All collected data in this dataset is precisely time-space aligned with the IMU measurements. Therefore, it is well-suited for experiments in time offset estimation\cite{qin2018}. This dataset enables evaluation of timing position error and overall VIO performance. The sequences are categorized based on motion dynamics: medium dynamic for the machine hall and high dynamic for the vicon room. To mimic real-case scenarios with time-varying time offsets, we introduce timing noise to the original timestamps for evaluating our online temporal calibration methods.
\par
\textit{Self-made Sensor-Cube (SCube) Dataset}:  For real-world sensor validation, we developed SCube, a multi-sensor device. This dataset comprises diverse locomotion types, such as handheld and vehicular movements, in indoor and outdoor settings. Angular velocity and linear acceleration were measured using an uncalibrated Xsens Mti-300-2A8G4 IMU at $200 \mathrm{~Hz}$. Visual images were collected using HIKROBOT MV-CA013-A0UC global-shutter cameras at $20\mathrm{~Hz}$. All sensors were triggered by the same clock source on a synchronization board. Indoor ground-truth positions were tracked using Opti-track, while outdoor ground-truth positions were recorded using an RTK GNSS receiver with a fixed solution. Perfect synchronization of all sensors was challenging due to trigger and data transmission latency. We employed Kalibr\cite{Rehder2016}, an open-source calibration tool, to estimate camera-IMU temporal offset offline, observing significant variations (up to 5 ms). This underscores the necessity for precise, robust online time offset estimation. Using SCube, we recorded 2 high dynamic motion indoor sequences with sharp maneuvers and 2 extensive outdoor sequences with continuous rotations for evaluation.
\begin{figure}
  \centering
  \includegraphics[width=9 cm]{"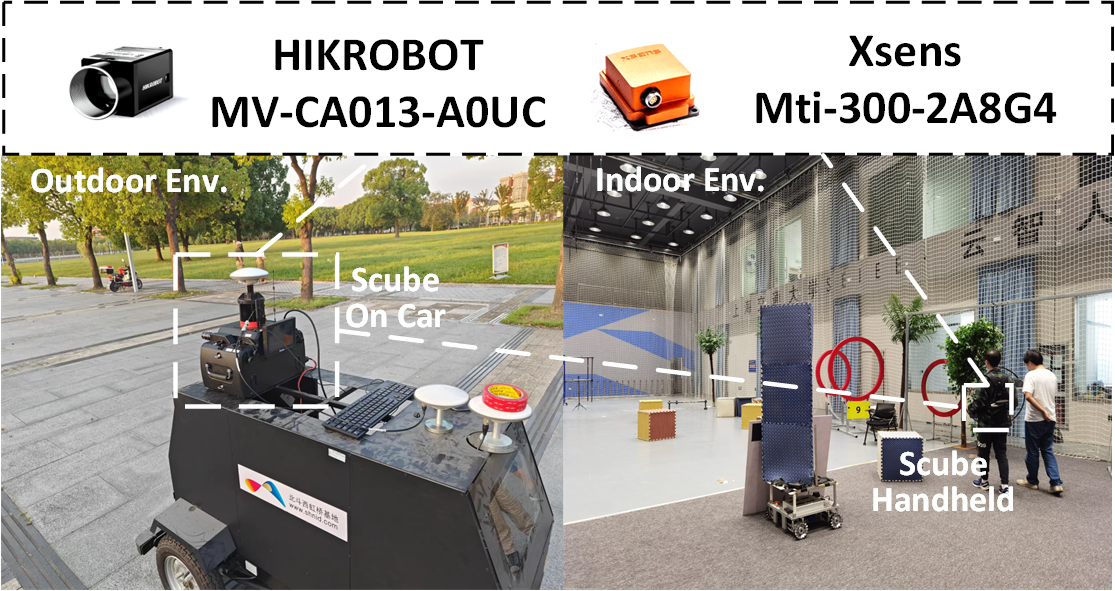"}
  \caption{The outlook of SCube and its application in outdoor and indoor environments. We utilized the camera and IMU to collect raw data with time-varying time offset.}
  \label{fig:Scube_APP}
\end{figure}
\begin{table}[htb!] \footnotesize
\renewcommand\arraystretch{1.2}
    \centering
    \begin{threeparttable}
    \caption{Datasets used for experimental evaluation.}
    \setlength{\tabcolsep}{2.5mm}{
    \begin{tabular}{|c|c|c|c|c|}
    \hline
    \textbf{Dataset} & \textbf{Duration[s] } & \textbf{Motion Dynamic Level}&\textbf{Vehicle} \\
    \hline
    AirSim & 74  & High & Car \\
    \hline
    EuRoC & 716  & Medium/High & Drone \\
    \hline
    Scube In. & 249  & High & Handheld \\
    \hline
    Scube Out. & 1700  & High & Car \\
    \hline
    \end{tabular}
    }
    \end{threeparttable}
\end{table}

\subsubsection{Compared Algorithms}
We compared our method against leading online temporal calibration approaches:
\begin{itemize}
\item VINS-Fusion \cite{qin2018}: An optimization-based VIO with conventional state-irrelevant online temporal calibration, denoted as VF\_SIR.
\item OpenVINS \cite{OpenVINS}: A filter-based VIO with state-relevant online temporal calibration, denoted as OV\_SR. Moreover, the conventional state-irrelevant method is implemented in this framework for comparison, denoted as OV\_SIR.
\end{itemize}

To ensure fairness for the comparison of temporal calibration performance in VIO, our proposed TON is integrated into the VINS-Fusion and OpenVINS under the same parameter settings without loop closure, denoted as VF\_TON and OV\_TON, respectively. The comparison is conducted only between the same VIO framework with identical visual tracking frontend and optimization backend.

Given the absence of test results for time-varying offsets in our datasets, we self-conducted these tests using modified open-source codes. All settings used for the compared algorithms will be publicly shared.

\subsection{Experimental Results and Discussions}
\subsubsection{Simulation}

In the simulation phase, our aim is to evaluate the convergence and accuracy of time offset estimation. Consequently, we set the time offset to various constant values on the collected two strictly aligned sequences AC01 and AC02 to demonstrate the ability for various large-range time offsets estimation. For sequence AC01, we set the time offset as -15ms, -10ms, +10ms, +15ms. For sequence AC02, we set the time offset as -20ms, -10ms, +10ms, +20ms. 
\par
\begin{figure*}[!t]
\centering
\subfloat[AirSim Dataset sequence AC01 with $t_d=15$ms]{\includegraphics[width=0.33 \textwidth]{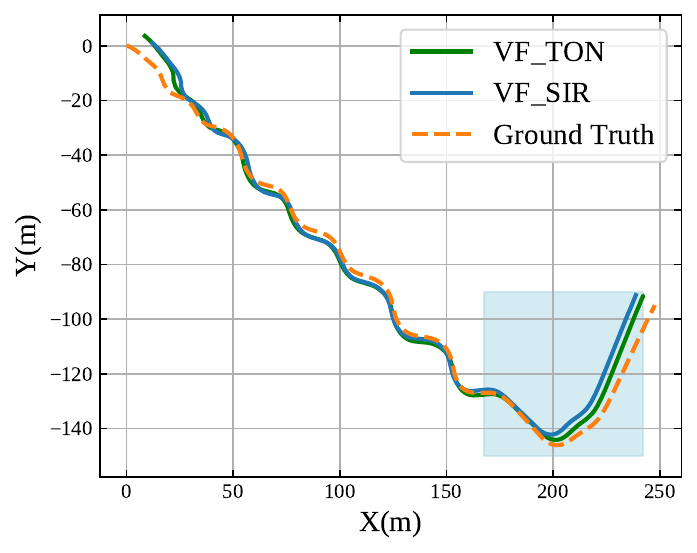}%
\label{fig_1_case}}
\hfil
\subfloat[High dynamic part of AC01 with $t_d=15$ms]{\includegraphics[width=0.33 \textwidth]{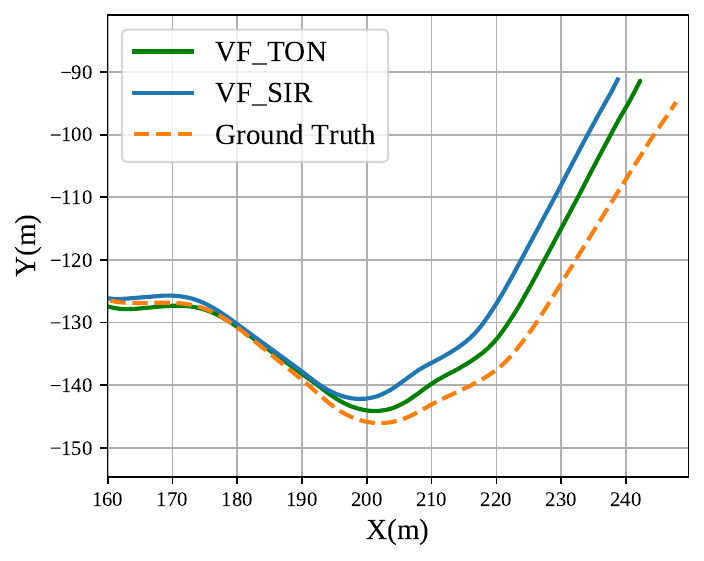}%
\label{fig_2_case}}
\subfloat[Time offset estimation error of AC01 with $t_d=15$ms]{\includegraphics[width=0.33 \textwidth]{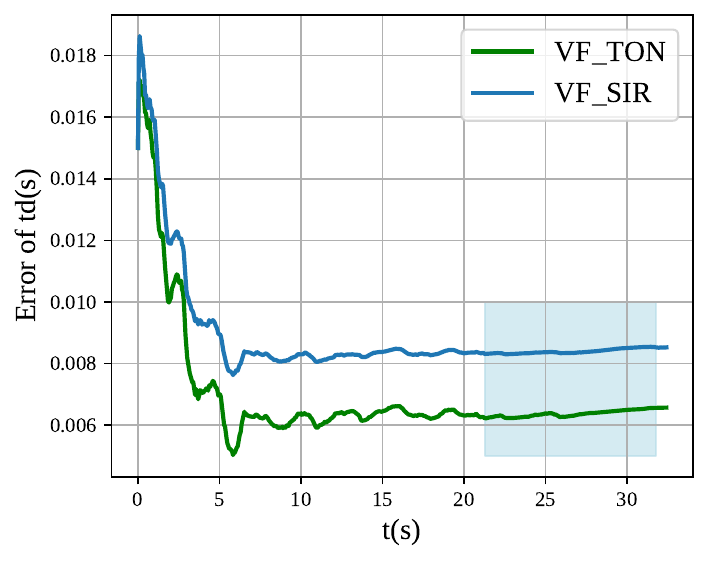}%
\label{fig_3_case}}
\caption{Comparison of trajectories and time offset estimation errors between proposed VF\_TON and baseline VF\_SIR on sequence AC01 with +15ms temporal misalignment. (a) shows the overall estimated trajectories of VF\_TON and VF\_SIR and highlights the extreme high-dynamic part. (b) enlarges the highlighted part in (a). (c) shows the estimation errors of time offset and highlights the estimating process corresponding to the highlighted part in (a).}
\label{fig_AC01}
\end{figure*}
\begin{figure*}[!t]
\centering
\subfloat[AirSim Dataset sequence AC02 with $t_d=10$ms]{\includegraphics[width=0.33 \textwidth]{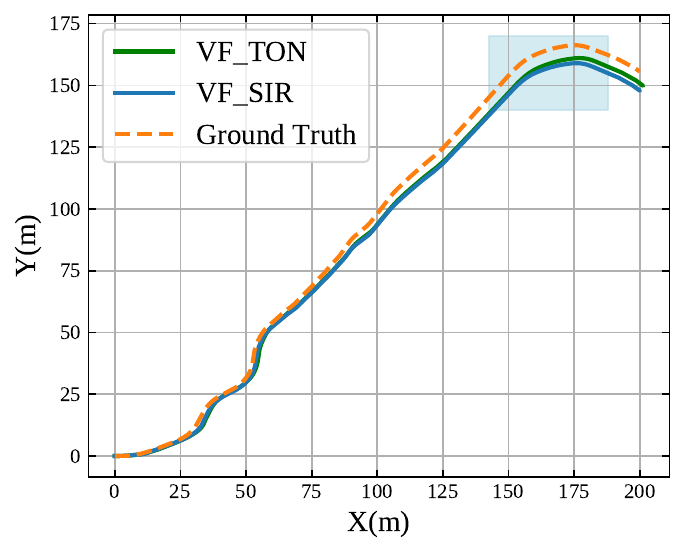}%
\label{fig_1_case}}
\hfil
\subfloat[High dynamic part of AC02 with $t_d=10$ms]{\includegraphics[width=0.33 \textwidth]{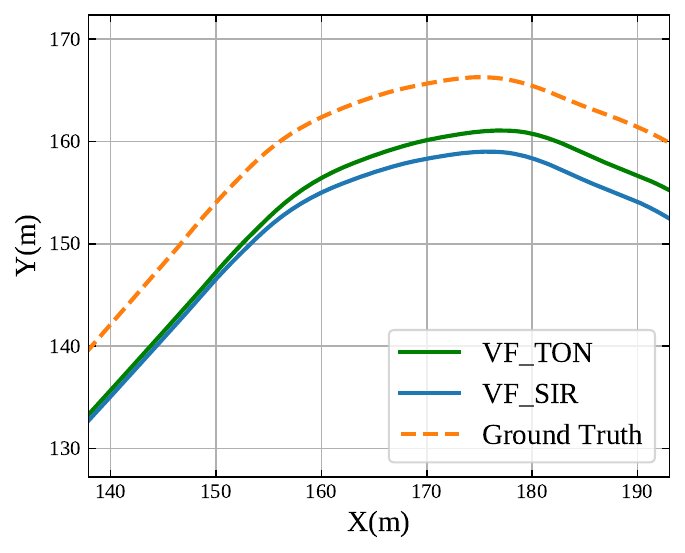}%
\label{fig_2_case}}
\subfloat[Time offset estimation error of AC02 with $t_d=10$ms]{\includegraphics[width=0.33 \textwidth]{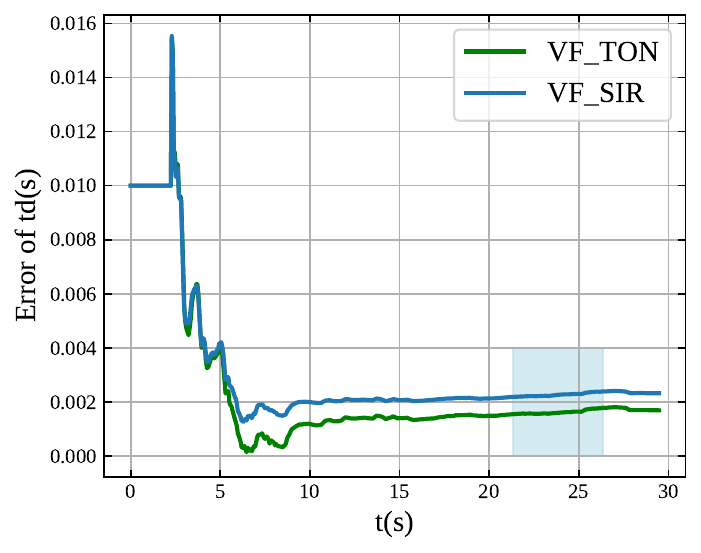}%
\label{fig_3_case}}
\caption{Comparison of trajectories and time offset estimation errors between proposed VF\_TON and baseline VF\_SIR on sequence AC02 with +10ms temporal misalignment. (a) shows the overall estimated trajectories of VF\_TON and VF\_SIR and highlights the extreme high-dynamic part. (b) enlarges the highlighted part in (a). (c) shows the estimation errors of time offset and highlights the estimating process corresponding to the highlighted part in (a).}
\label{fig_AC02}
\end{figure*}

\textit{Time Offset Convergence Evaluation}:
 For the VINS-Fusion based algorithms, our proposed VF\_TON achieves lower CIT compared with the original VF\_SIR in all AirSim sequences, which indicates the faster convergence of our method. As depicted in Fig. \ref{fig_AC01} (c) and Fig. \ref{fig_AC02}, the time offset estimation accuracy of our VF\_TON is also superior to VF\_SIR, which eventually improves the positioning performance.

OpenVINS based algorithms fail in sequence AC01 due to bad initialization and fragile state estimation, which is not related to the temporal calibration. So the evaluation is only conducted in sequence AC02. Although the CIT of OV\_SR and OV\_SIR seems to be comparable to OV\_TON, the VIO overall performance is worse than our OV\_TON. The reason is that the temporal calibration methods of OV\_SR and OV\_SIR have more negative impact on state estimation due to inaccurate observation and prediction modeling for the time offset. 

\begin{table*}[htb] \footnotesize
\renewcommand\arraystretch{1.1}
    \centering
    \caption{AirSim: CIT, Root Mean Square (RMSE) of Absolute Position Error (APE)[M] and Rotation Error (ARE)[Degree]}
    \begin{tabular}{|l|c|c|c|c|c|c|c|c|c|c|c|c|c|c|c|c|}
    \hline
     &  &  \multicolumn{3}{c|}{VF\_SIR} & \multicolumn{3}{c|}{\textbf{VF\_TON}} & \multicolumn{3}{c|}{OV\_SR} & \multicolumn{3}{c|}{OV\_SIR} & \multicolumn{3}{c|}{\textbf{OV\_TON}} \\
     \hline
    Seq. & td[ms]& CIT & APE & ARE & CIT & APE & ARE & CIT & APE & ARE & CIT & APE & ARE & CIT & APE & ARE \\
    \hline \multirow{4}{*}{AC01} & -15 &  56 & 3.19 & 5.88 & \textbf{46} &  \textbf{3.02 }& \textbf{5.37 }& - & -& -& -& -& -& -& -& - \\
                                 & -10  &  52 & 3.43 & 6.05 &  \textbf{37} & \textbf{3.15} &\textbf{5.54 }& - & -& -& -& -& -& -& -& -  \\
                                 & +10  &  -  & 4.32 & 5.75 &  \textbf{88}  & \textbf{3.77} & 5.93 & - & -& -& -& -& -& -& -& -  \\
                                 & +15  &  - & 5.76 & 5.95 &  \textbf{96} & \textbf{3.68} & 6.18 & - & -& -& -& -& -& -& -& -  \\
    \hline \multirow{4}{*}{AC02} 
                                 & -20  & 74 & 1.57 & 2.02 &  \textbf{61} & 1.72 & 2.10 & 54 & 3.39 & 7.17 & 54 & 3.34 & 6.77 & 54 & \textbf{3.23} & \textbf{6.28} \\
                                 & -10  & 43 & 1.94 & 1.79 & \textbf{43}& \textbf{1.68}& \textbf{1.79} & \textbf{44} & 3.43 & 6.51 & 45 & 3.30 & 6.78 & 78 & \textbf{3.07} &  \textbf{5.48} \\
                                 & +10 & - & 2.46 & 2.26 & \textbf{103} & \textbf{1.91 }& \textbf{1.81 }& 44 & 3.57 & 7.60 & 76 & 3.37 & 6.32 & \textbf{33} & \textbf{3.36} & 6.63 \\
                                 & +20 & - & 2.47 & 2.13 & \textbf{144 }&\textbf{1.78} & \textbf{1.81} & 94 & 3.57 & 7.67 & 78 & 3.70 & 6.21 & \textbf{36} & \textbf{2.36} & \textbf{3.10} \\
    \hline
    \end{tabular}
\label{table_AC}
\end{table*}

\textit{VIO Overall Performance}:
Next, the positioning performance of VIO with our proposed TON is evaluated. As shown in Table \ref{table_AC}, our approach demonstrates superior stability and accuracy. For nearly all AirSim sequences with different time offsets, the positioning accuracy of TON enhanced VIO systems outperform their baselines. Moreover, the positioning stability of our TON surpass the baselines under different time offsets, indicating that the proposed TON has stronger ability to robustly estimate various large-range time offsets.

The estimated trajectories of the algorithms are depicted in Fig. \ref{fig_AC01} (a) and Fig. \ref{fig_AC02} (a). The conventional state-irrelevant method VF\_SIR struggles to estimate the time offset under motions with high linear and angular velocity because of unstable visual tracking conditions. In contrast, our proposed TON can effectively handle these motions with more accurate observation and prediction model of the time offset.


\subsubsection{EuRoC Dataset}
The primary objective of evaluation on EuRoC Dataset is to validate the capability of online temporal calibration methods in estimating time-varying time offsets and evaluating the observation errors resulting from these offsets. To generate time-varying time offset, two different cumulative biases and white Gaussian noises are added on every tested sequence.

\par
\begin{figure*}[!t]
\centering
\subfloat[APE of OpenVINS based algorithms on EuRoC sequence V1\_03\_H]{\includegraphics[width=0.5 \textwidth]{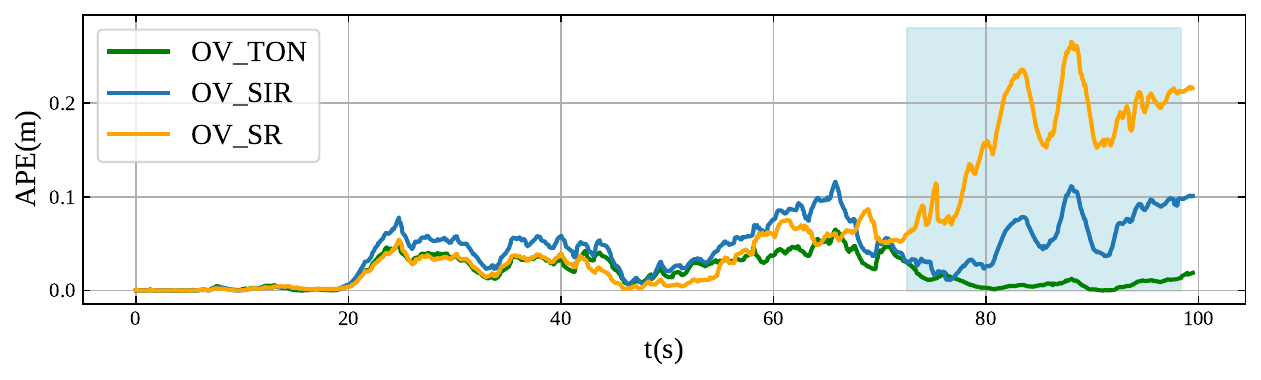}%
\label{fig_1_case}}
\subfloat[TPE of OpenVINS based algorithms on EuRoC sequence V1\_03\_H]{\includegraphics[width=0.5 \textwidth]{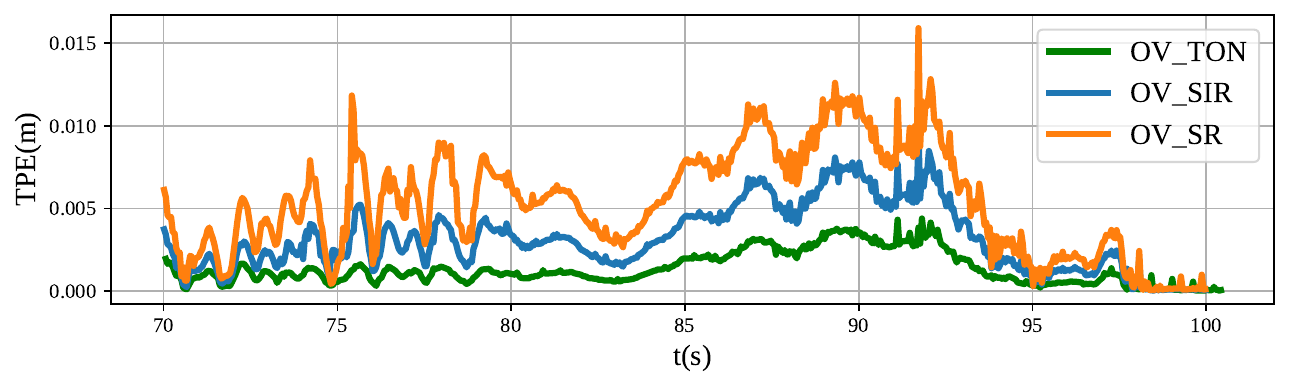}%
\label{fig_2_case}}
\hfil
\subfloat[APE of VINS-Fusion based algorithms on EuRoC sequence V1\_03\_H]{\includegraphics[width=0.5 \textwidth]{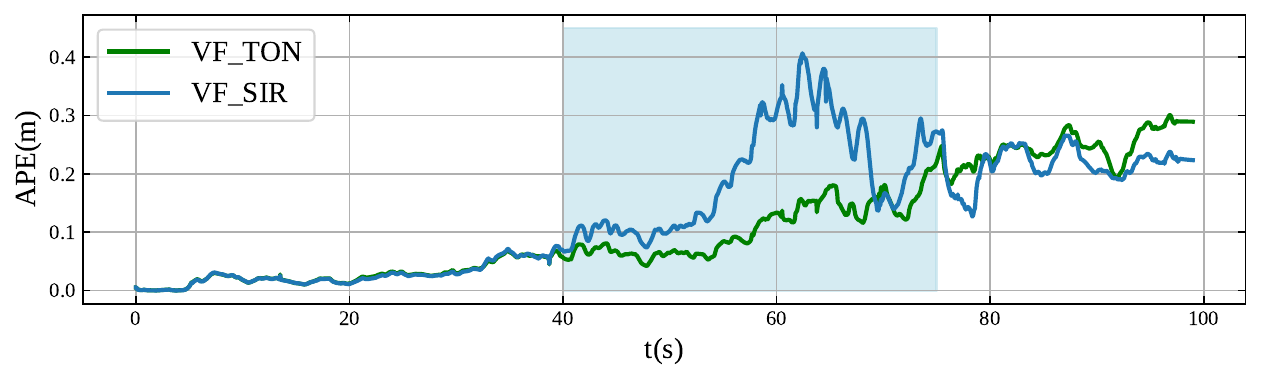}%
\label{fig_3_case}}
\subfloat[TPE of VINS-Fusion based algorithms on EuRoC sequence V1\_03\_H]{\includegraphics[width=0.5 \textwidth]{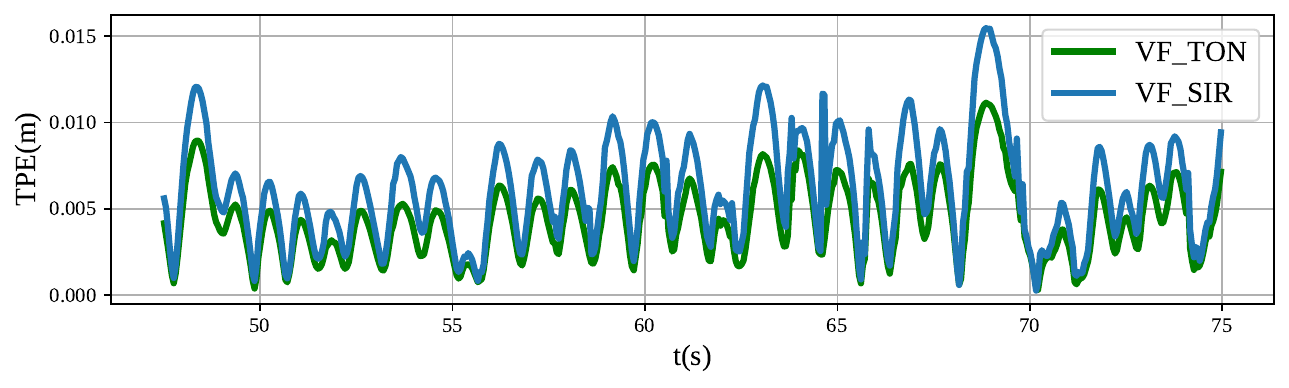}%
\label{fig_4_case}}
\caption{Comparison of APE and TPE between VIO systems with and without our proposed TON on EuRoC sequence V1\_03\_H with $n_{t_d}=0.1$ms cumulative temporal misalignment. (a) shows APE of OpenVINS based algorithms and highlights the significant drifting part. (b) shows TON corresponding to the highlighted part in (a). (c) shows APE of VINS-Fusion based algorithms and highlights the significant drifting part. (d) shows TON corresponding to the highlighted part in (c).}
\label{fig_V103}
\end{figure*}
\begin{figure*}[!t]
\centering
\subfloat[APE of OpenVINS based algorithms on EuRoC sequence V2\_02\_H]{\includegraphics[width=0.5 \textwidth]{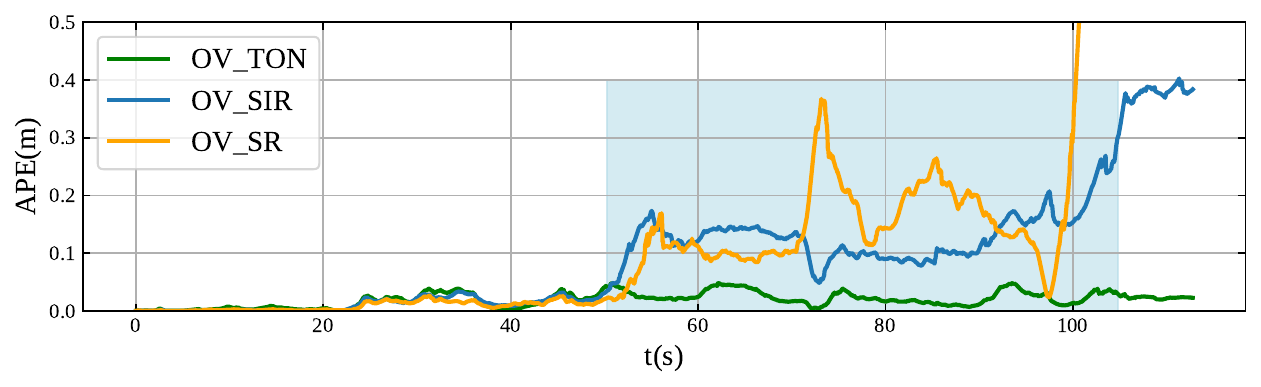}%
\label{fig_1_case}}
\subfloat[TPE of OpenVINS based algorithms on EuRoC sequence V2\_02\_H]{\includegraphics[width=0.5 \textwidth]{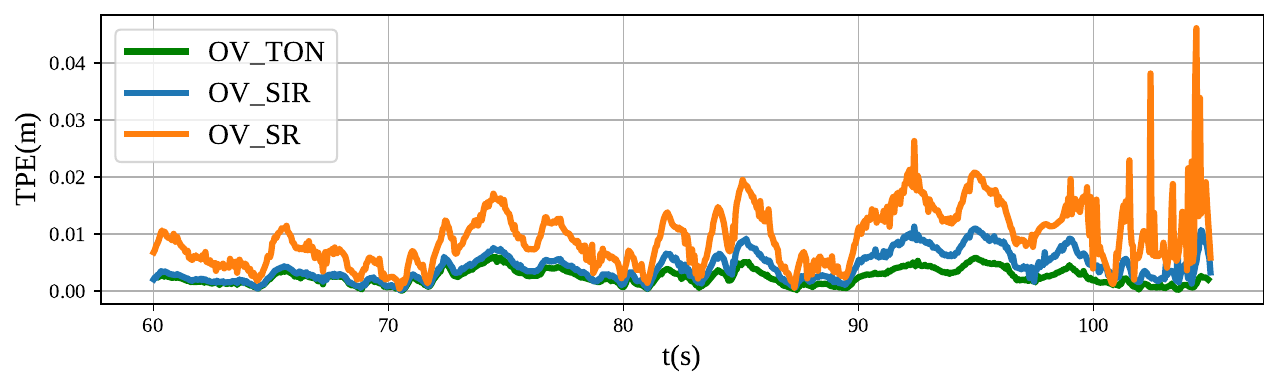}%
\label{fig_2_case}}
\hfil
\subfloat[APE of VINS-Fusion based algorithms on EuRoC sequence V2\_02\_H]{\includegraphics[width=0.5 \textwidth]{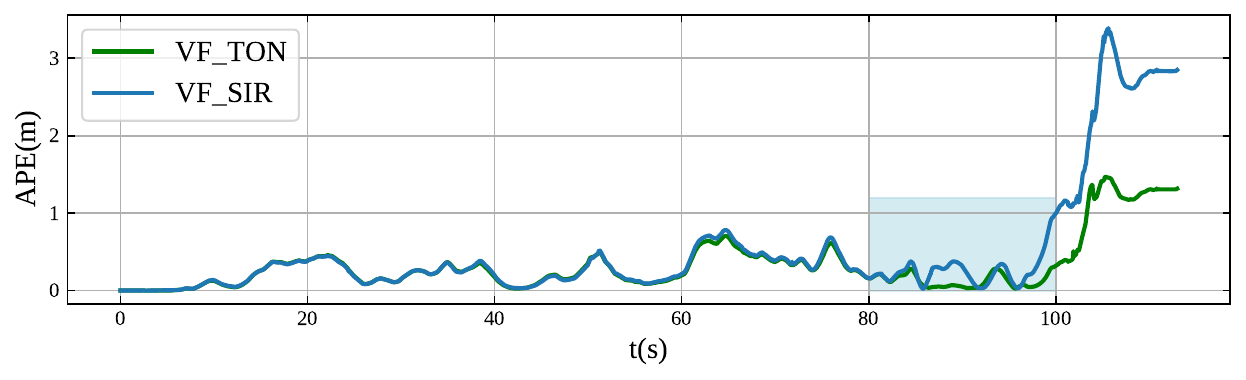}%
\label{fig_3_case}}
\subfloat[TPE of VINS-Fusion based algorithms on EuRoC sequence V2\_02\_H]{\includegraphics[width=0.5 \textwidth]{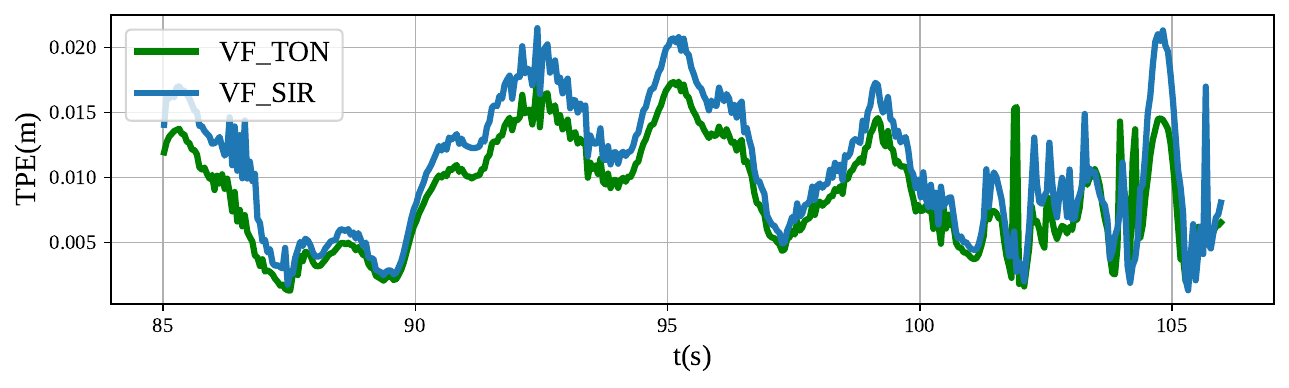}%
\label{fig_4_case}}
\caption{Comparison of APE and TPE between VIO systems with and without our proposed TON on EuRoC sequence V2\_02\_H with $n_{t_d}=0.1$ms cumulative temporal misalignment. (a) shows APE of OpenVINS based algorithms and highlights the significant drifting part. (b) shows TON corresponding to the highlighted part in (a). (c) shows APE of VINS-Fusion based algorithms and highlights the significant drifting part. (d) shows TON corresponding to the highlighted part in (c).}
\label{fig_V202}
\end{figure*}

\textit{Timing Error Evaluation}:
 Table \ref{table_EuRoC} displays the TPE for all medium and high dynamic motion sequences in the EuRoC dataset. For the VINS-Fusion based algorithms, in almost all sequences, our proposed VF\_TON improves TPE by an average of 19\%. When compared to OpenVINS based methods, our OV\_TON achieves lower TPE than the OV\_SR by an average of 46\%. Furthermore, as depicted in Fig. \ref{fig_V103} and Fig. \ref{fig_V202}, the improvement 
 of TPE by our proposed VF\_TON and OV\_TON become especially significant in high dynamic motions in the challenging vicon room sequences, indicating superior performance in high dynamic motions.

\begin{table*}[htb!] \footnotesize

\renewcommand\arraystretch{1.4}
    \centering
    \begin{threeparttable}
    \caption{EuRoC: RMSE of APE[M] and TPE[mm]}
    \begin{tabular}{|c|c|c|c|c|c|c|c|c|c|c|c|c|c|c|c|}
    \hline
      \multicolumn{2}{|c|}{Sequence}  & \multicolumn{2}{c|}{MH\_03\_M} &  \multicolumn{2}{c|}{MH\_04\_M}  &  \multicolumn{2}{c|}{MH\_05\_M} &  \multicolumn{2}{c|}{V1\_02\_H} &  \multicolumn{2}{c|}{V1\_03\_H} &  \multicolumn{2}{c|}{V2\_02\_H} &  \multicolumn{2}{c|}{V2\_03\_H}     \\
      \hline
      $n_{t_d}^{1}$[ms] & Methods & TPE & APE  & TPE & APE & TPE & APE & TPE & APE & TPE & APE & TPE & APE & TPE & APE  \\
    \hline \multirow{5}{*}{$0.1$}  & OV\_SR & 8.8 & 0.118 &  5.6& 0.219& 6.3& 0.249& 4.7& 0.113& 4.4& 0.227& 7.4& 0.606& 6.0& 1.045\\
    & OV\_SIR & 7.0 & 0.122& 3.6& 0.201& 5.0& 0.261& 3.1& 0.104& 2.9& 0.166& 3.7& 0.201& 3.3& 0.528\\
    & \textbf{OV\_TON} & \textbf{5.6}& \textbf{0.101} & \textbf{2.7}& \textbf{0.138} & \textbf{4.5}& \textbf{0.191} & \textbf{1.9}& \textbf{0.090} & \textbf{2.7}& \textbf{0.095} & \textbf{2.7}& \textbf{0.078} & \textbf{2.3}& \textbf{0.355} \\
    \cline{2-16}
    & VF\_SIR & 10.2 & 0.210 & 7.0 & 0.452 & 8.5 & 0.298 & 6.2 & 0.155 & 6.1 & 0.286 & 7.0 & 0.468 & 6.6 & 0.656 \\
    & \textbf{VF\_TON} & \textbf{8.9} & \textbf{0.190} & \textbf{5.8} & \textbf{0.447} & \textbf{6.9} & \textbf{0.286} & \textbf{5.2} & \textbf{0.140} & \textbf{4.7} & \textbf{0.209} & \textbf{5.8} & \textbf{0.274} & \textbf{5.8} & \textbf{0.521} \\
    \hline
    \multirow{5}{*}{$0.125$}  & OV\_SR & 11.7& 0.144& 7.0& 0.215& 7.7& 0.255& 6.0& 0.127& 5.7& 0.267& 9.6& 1.164& 7.9& 0.998\\
    & OV\_SIR & 9.4& 0.153& 4.3& 0.214& 5.8& 0.256& 3.9& 0.109& 4.0& 0.213& 4.5& 0.294& 5.0& 0.778\\
    & \textbf{OV\_TON} & \textbf{7.1}& \textbf{0.127} & 4.8& \textbf{0.128} & \textbf{4.3}& \textbf{0.215} & \textbf{3.5}& \textbf{0.093} & 4.0& \textbf{0.117} & \textbf{3.3}& \textbf{0.079} & \textbf{3.6}& \textbf{0.592} \\
    \cline{2-16}
    & VF\_SIR & 8.0 & 0.224 & 5.4 & 0.448 & 6.4 & 0.332 & 5.7 & 0.164 & 5.7 & 0.419 & 6.6 & 0.415 & {6.3} & 1.009 \\
    & \textbf{VF\_TON} & \textbf{6.4} & \textbf{0.201} & \textbf{4.4} & 0.500 & \textbf{5.5} & \textbf{0.328} & \textbf{4.5} & \textbf{0.158} & \textbf{3.7} & \textbf{0.258} & \textbf{5.1} & \textbf{0.338} & \textbf{5.2} & \textbf{0.636} \\
    \hline
    \end{tabular}
    \begin{tablenotes}
    \item[1] The timing noise represents the accumulative time-varying time offset drifting noise per second.
    \end{tablenotes}
    \end{threeparttable}
\label{table_EuRoC}
\end{table*}

\begin{table}[htb] \footnotesize
    \centering
    \renewcommand\arraystretch{1.1}
    \begin{threeparttable}
    \caption{Scube: RMSE and Standard Deviation (STD) of APE[M] in Ablation Study}
    \setlength{\tabcolsep}{1.5mm}{
    \begin{tabular}{|c|c|c|c|c|c|c|c|c|}
    \hline Seq. & \multicolumn{2}{c|}{VF\_SIR} & \multicolumn{2}{c|}{VF\_FVON} & \multicolumn{2}{c|}{VF\_TPN} & \multicolumn{2}{c|}{\textbf{VF\_TON}} \\
    \hline 
    & RMSE & STD & RMSE & STD & RMSE & STD & RMSE & STD \\
    \hline \multirow{1}{*}{SI01} & 1.01 & 0.60 & 0.49 & 0.20 & 0.74 & 0.36 & \textbf{0.46} & \textbf{0.17}   \\
           \multirow{1}{*}{SI02} & 0.74 & 0.34 & 0.73 & 0.34 & 0.71 & \textbf{0.23} & \textbf{0.65} & 0.25   \\
           \multirow{1}{*}{SO01} & 5.91 & 2.68 & 4.90 & 2.24 & 5.20 & 2.46 & \textbf{4.32} & \textbf{1.79}  \\
           \multirow{1}{*}{SO02} & 19.72 & 10.38 & 14.21 & 7.77 &  19.90 & 10.74 &  \textbf{13.22} & \textbf{5.37}\\
    \hline
    \end{tabular}
    }
    \end{threeparttable}
\label{table_scube}
\end{table}

\textit{VIO Overall Performance}: As indicated in Table IV, our proposed VF\_TON and OV\_TON demonstrates a notable reduction in positioning error. VF\_TON outperforms VF\_SIR by an average of 15\%, while OV\_TON surpasses OV\_SR by an average of 42\% across all test sequences. The improvement in positioning error is highly consistent to TPE, which proves that our proposed TON enhances the VIO positioning performance by superior accuracy of time offset estimation. Our method's efficacy becomes increasingly apparent with higher motion speeds, especially in rotational movements within the vicon room sequences. This underlines our method's enhanced robustness and effectiveness in time offset estimation for better pose accuracy compared to both conventional state-relevant and state-irrelevant methods.

\subsubsection{Self-Collected Scube Dataset}
We collected raw data facilitating real sensor sets to evaluate the performance of our proposed TON in real-world scenarios. The timestamps of these self-collected sequences are kept original, without any post-processing. Furthermore, ablation studies of our proposed FVON and TPN in the TON are conducted to show the effectiveness of the designed observation and prediction modules. Note that OpenVINS based algorithms are too fragile to run on the Scube Dataset so that we do not introduce them for comparisons in this part of experiment.

\begin{figure*}[!t]
\centering
\subfloat[Scube Dataset sequence SI01]{\includegraphics[width=0.25 \textwidth]{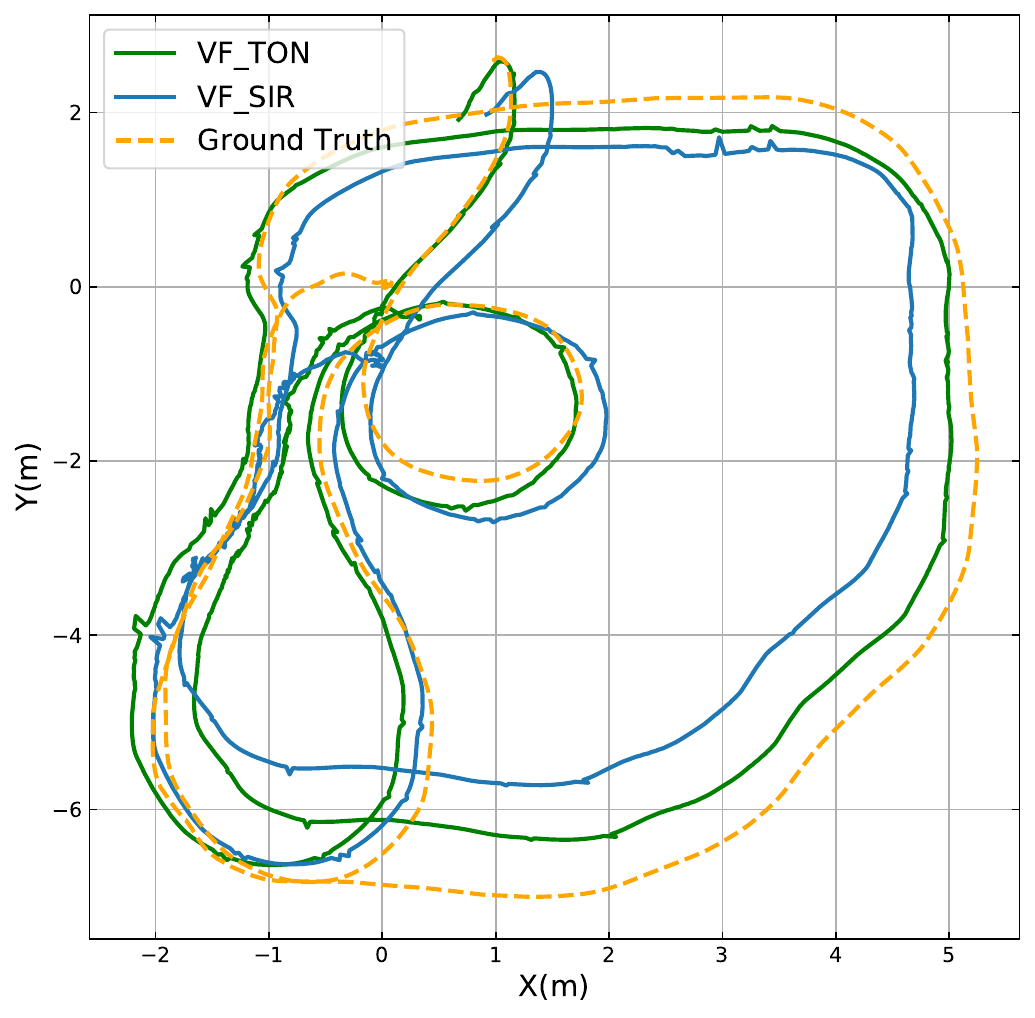}%
\label{fig_1_case}}
\subfloat[Scube Dataset sequence SI02]{\includegraphics[width=0.25 \textwidth]{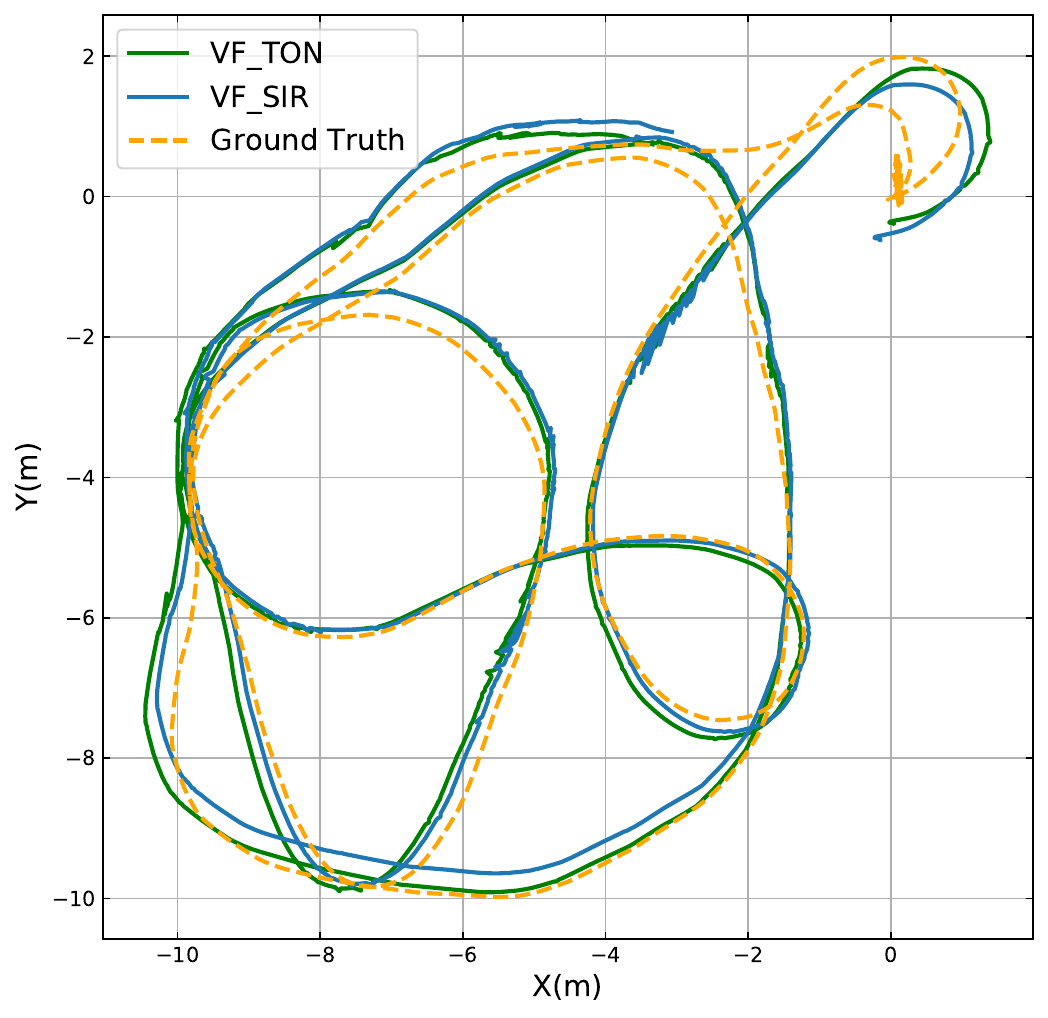}%
\label{fig_2_case}}
\subfloat[Scube Dataset sequence SO01]{\includegraphics[width=0.25 \textwidth]{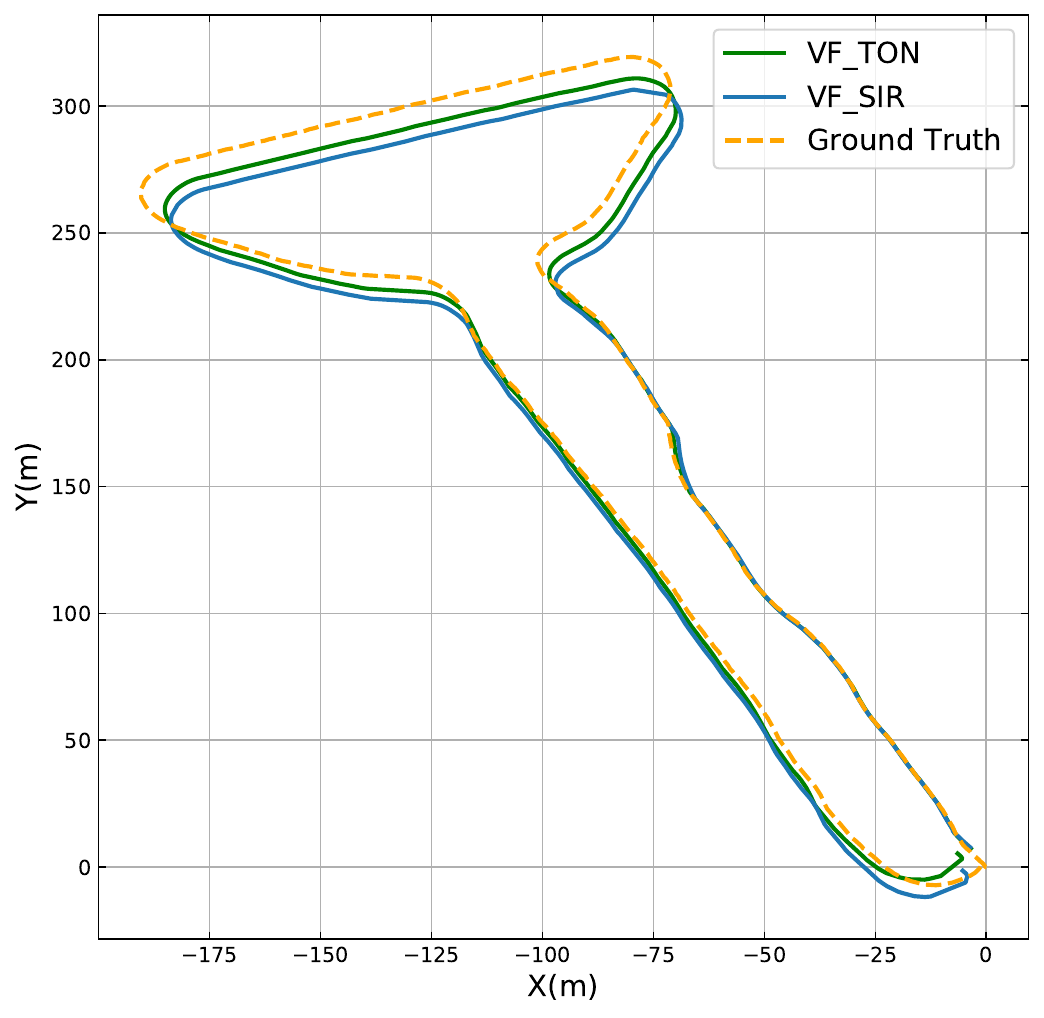}%
\label{fig_3_case}}
\subfloat[Scube Dataset sequence SO02]{\includegraphics[width=0.25 \textwidth]{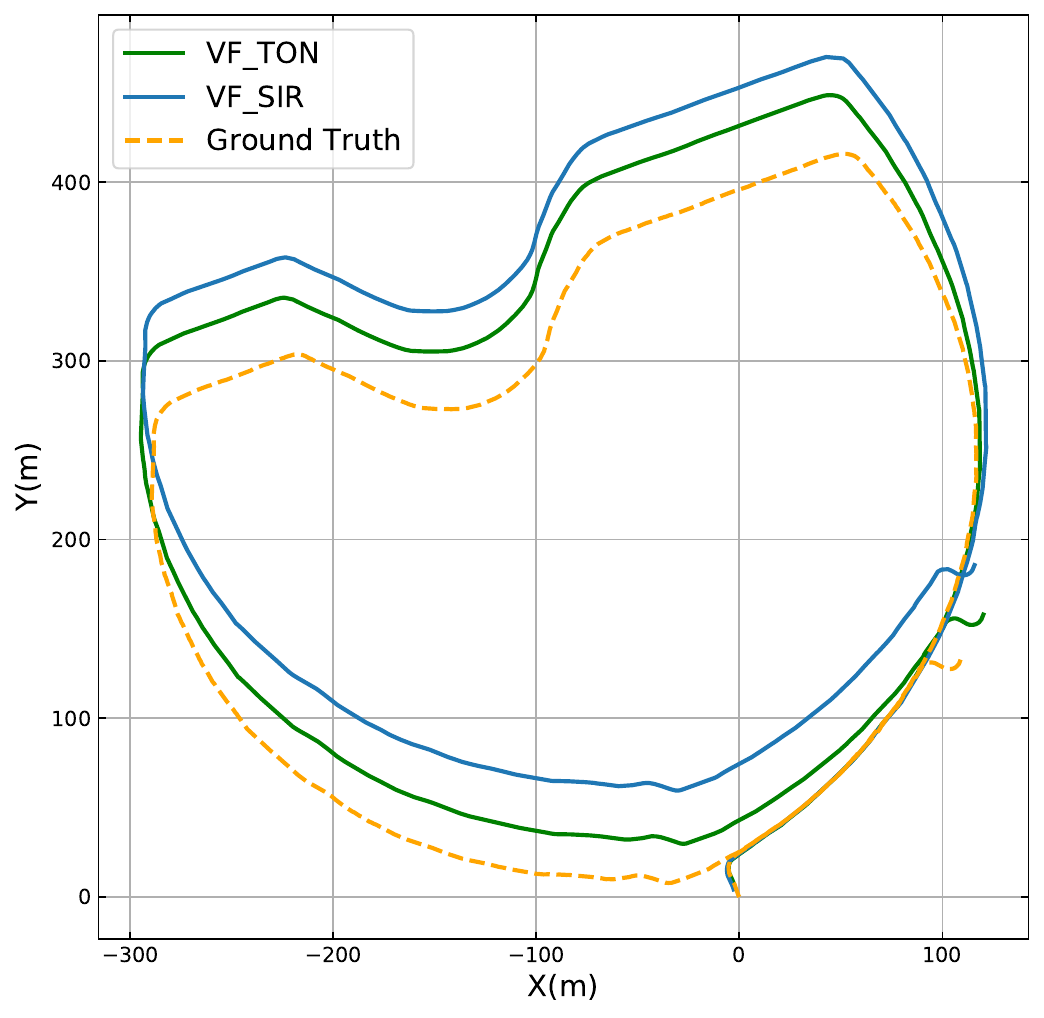}%
\label{fig_3_case}}
\caption{Comparison of trajectories between proposed VF\_TON and VF\_SIR on Scube sequences with raw timestamps.}
\label{fig_scube}
\end{figure*}

\par
\textit
{VIO Overall Performance}: To showcase our method's effectiveness, we assess the APE across all sequences. 
Table \ref{table_scube} shows the APE for all Scube dataset sequences. Our method achieves an average reduction in APE of 32\% for all tested sequences. The indoor tests are characterized by sharp turns, as shown in Fig. \ref{fig_scube} (a) and (b), while the outdoor sequences feature prolonged rotational movements, as shown in Fig. \ref{fig_scube} (c) and (d). These findings underscore our method's superior effectiveness and robustness compared to the conventional method, particularly in high dynamic motion scenarios with real sensor setups.

\textit
{Ablation Study}:
The two modules of our proposed TON, FVON and TPN, are independently implemented in VINS-Fusion for ablation study, denoted as VF\_FVON and VF\_TPN, respectively. As shown in Table V, VF\_FVON reduces the APE compared to conventional VF\_SIR, while VF\_TPN seems to have less improvement. The reason is that without our proposed FVON, the estimated time offset is less accurate, so that the proposed TPN are not able to learn the evolution pattern of time offset effectively. Furthermore, the combination of the two modules VF\_TON has the overall best performance. These results indicate the effectiveness of our designed modules for observation and prediction model of time offset.

\section{Conclusions}
In this paper, we propose novel online weakly-supervised learning networks TON for online temporal calibration. TON is composed of FVON for time offset observation modeling and TPN for prediction modeling. The proposed FVON can effectively compute the velocity of poorly tracked features' velocity utilized in the time offset estimation constraints to improve the observation modeling. The proposed TPN can predict the time offset by learning from historical evolution, enhancing the prediction modeling. Integrating our proposed FVON and TPN into state estimators in both optimization-based and filter-based VIO frameworks can improve the time offset estimation in long-term or high dynamic motion challenging scenarios.
\par
Experiments are conducted to compare our TON with other leading approaches. Initially, we use AirSim to create a strictly synchronized dataset, enabling an effective evaluation of our time offset estimation method. Subsequently, we test our approach on the EuRoC dataset and utilize our self-made device, SCube, for gathering real-world indoor and outdoor data. The results indicate that our TON achieves more robust time offset estimation in long-term and high dynamic motion scenarios.
\par
Although we have showcased the integration of our method in two VIO frameworks VINS-Fusion and OpenVINS in this paper, the proposed TON can be generalized to other VIO systems for more accurate time offset modeling in online temporal calibration.

\normalem


 

\begin{IEEEbiography}[{\includegraphics[width=1in,height=1.25in,clip,keepaspectratio]{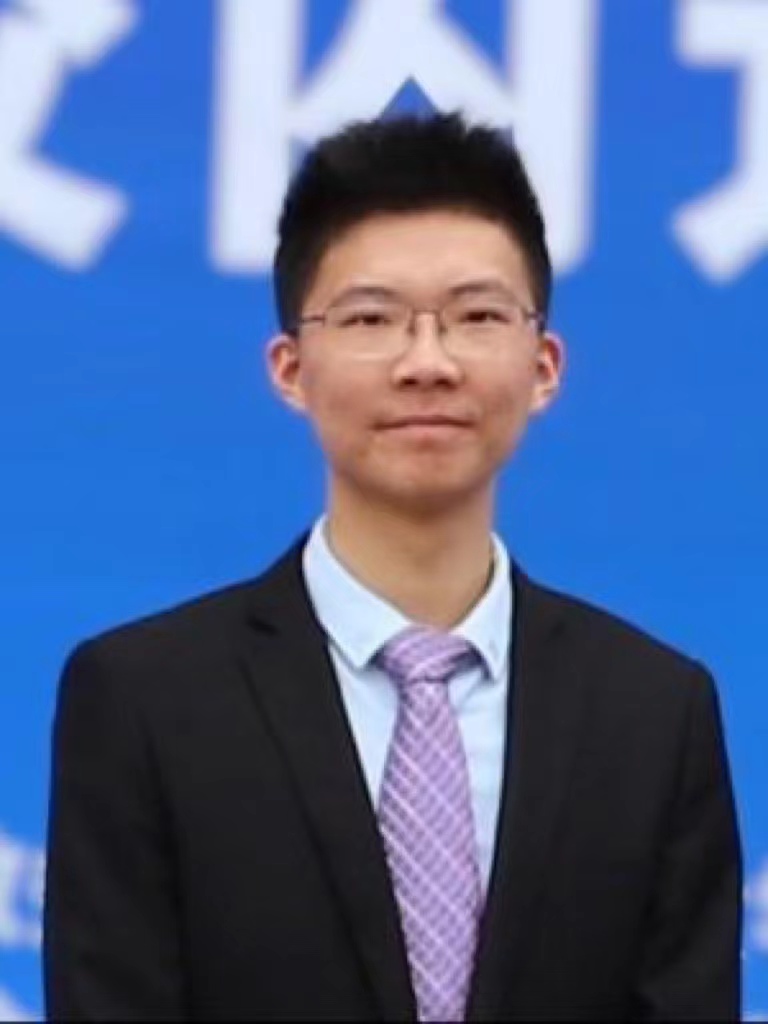}}]{Chaoran Xiong}
(Student Member, IEEE) received the B.S. degree in School of Electronic and Information Engineering, Soochow University, Suzhou, China, in 2023. He is currently pursuing the Ph.D. degree with Shanghai Jiao Tong University, Shanghai, China. His current research interests include multi-modal interactive sensing and embodied navigation.
\end{IEEEbiography}

\begin{IEEEbiography}[{\includegraphics[width=1in,height=1.25in,clip,keepaspectratio]{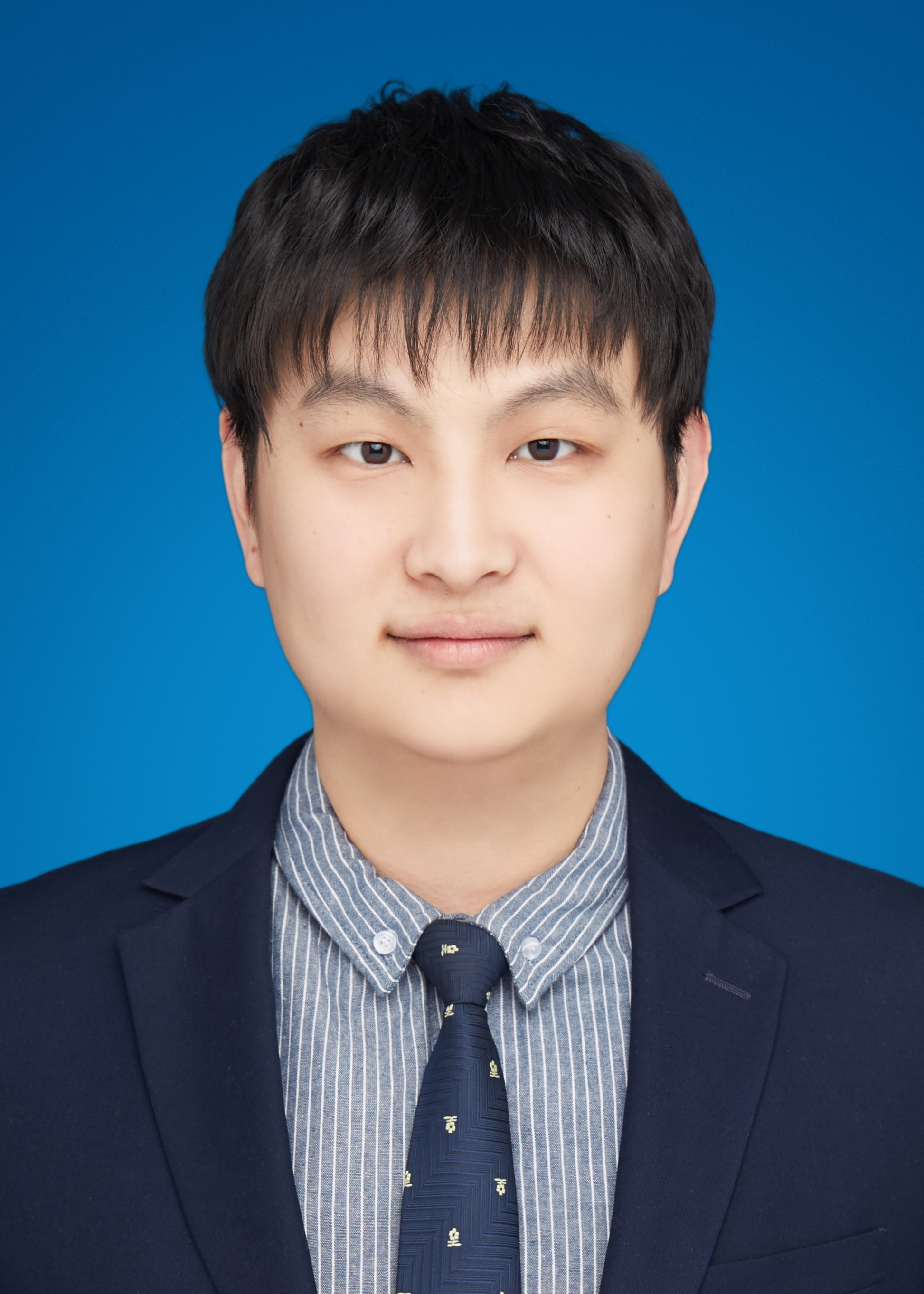}}]{Guoqing Liu}
received the bachelor’s degree in automation from the School of Automation, Harbin University of Science and Technology,
Harbin, China, in 2018, and the master’s degree in robotics science and engineering with the Faculty of Robot Science and Engineering, North-eastern University, Shenyang, China, in 2021. 
He is currently pursuing the Ph.D. degree with Shanghai Jiao Tong University, Shanghai, China.
\end{IEEEbiography}

\begin{IEEEbiography}[{\includegraphics[width=0.95in,height=1.2in,clip]{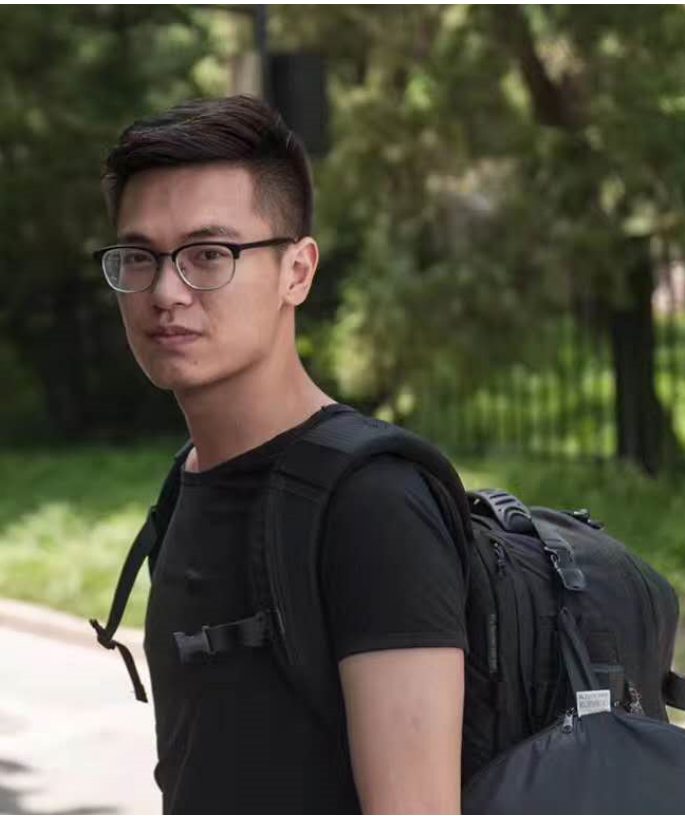}}]{Qi Wu}
(Student Member, IEEE) received the bachelor’s degree in electrical engineering and
automation from the Chongqing University of
Posts and Telecommunications, Chongqing,
China, in 2016, and the master’s degree in electronic science and technology from the Beijing
University of Posts and Telecommunications,
Beijing, China, in 2019. He is currently working
toward the Ph.D. degree in electronic science
and technology with Shanghai Jiao Tong University, Shanghai, China.
\end{IEEEbiography}

\begin{IEEEbiography}[{\includegraphics[width=1in,height=1.25in,clip,keepaspectratio]{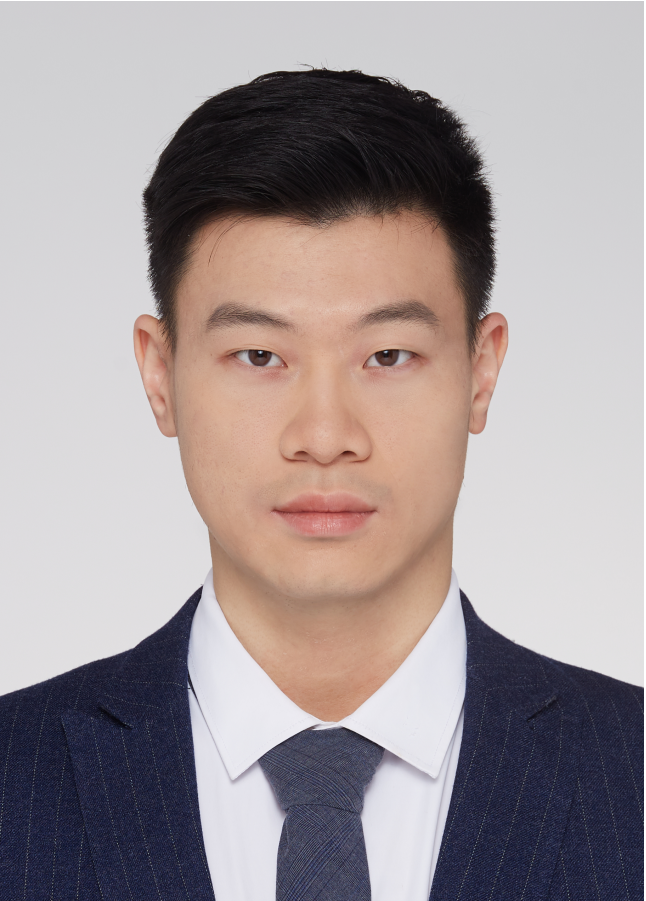}}]{Songpengcheng Xia}
(Graduate Member. IEEE)
received the B.S. degree in navigation engineering from Wuhan University, Wuhan, China,
in 2019. He is currently pursuing the Ph.D.
degree with Shanghai Jiao Tong University,
Shanghai, China. His current research interests
include machine learning, inertial navigation,
multi-sensor fusion and wearable-based human
activity recognition.
\end{IEEEbiography}

\begin{IEEEbiography}[{\includegraphics[width=1in,height=1.25in,clip,keepaspectratio]{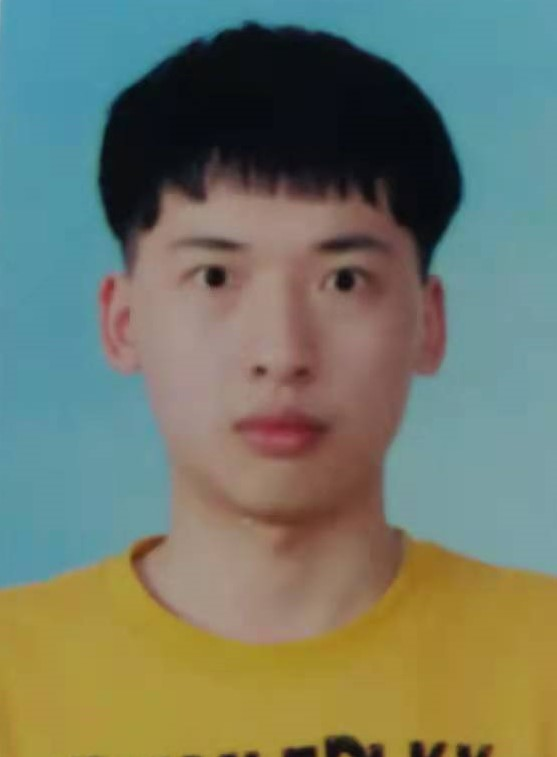}}]{Tong Hua}
received the B.S. degree in information engineering from Shanghai Jiao Tong University, Shanghai, China, in 2021. He is currently pursuing a M.S. degree at Shanghai Jiao Tong University, Shanghai, China. His current research interests include Visual-Inertial SLAM, multi-sensor fusion.
\end{IEEEbiography}

\begin{IEEEbiography}[{\includegraphics[width=1in,height=1.25in,clip,keepaspectratio]{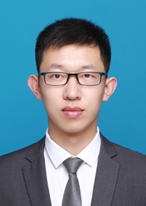}}]{Kehui Ma}
received the B.S. degree in Intelligent Science and Technology from Xidian University, Xi'an, China, in 2022. He is currently pursuing the Ph.D. degree with Shanghai Jiao Tong University, Shanghai, China. His current research interests include reinforcement learning and embodied navigation.
\end{IEEEbiography}

\begin{IEEEbiography}[{\includegraphics[width=1in,height=1.25in,clip,keepaspectratio]{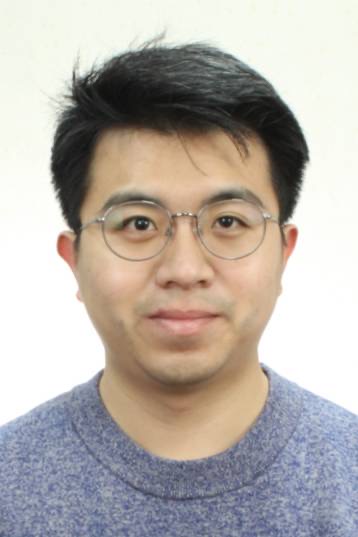}}]{Zhen Sun}
received his B.S. degree from Harbin Institute of Technology in 2017 and M.S. degree from Shanghai Jiao Tong University in 2020, and has been pursuing his Ph.D. degree at Shanghai Jiao Tong University since then. His current research interests are in the research of theoretical methods of brain-inspired navigation for unknown and complex environments.
\end{IEEEbiography}

\begin{IEEEbiography}[{\includegraphics[width=1in,height=1.25in,clip,keepaspectratio]{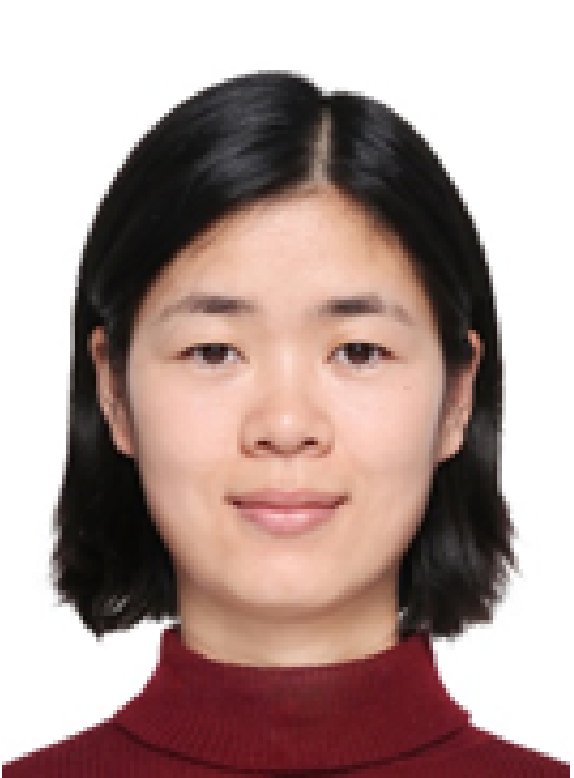}}]{Yan Xiang}
received her Ph. D. in the Department of Geomatics Engineering at University of Calgary, in 2018. She is an accociate professor at Shanghai Jiao Tong University. Her current research interests include GNSS inter-frequency and inter-system code and phase bias determination, carrier-phase based ionospheric modeling for GNSS high precision positioning, and multi-sensor integration.
\end{IEEEbiography}

\begin{IEEEbiography}[{\includegraphics[width=1in,height=1.25in,clip,keepaspectratio]{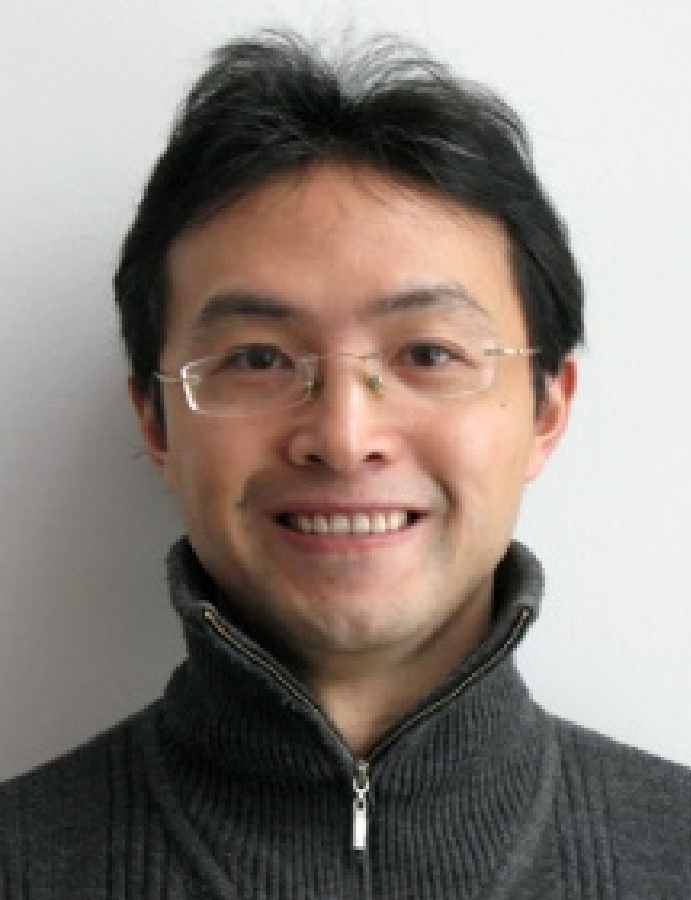}}]{Ling Pei}
(Senior Member, IEEE) received the
Ph.D. degree from Southeast University, Nanjing, China, in 2007. From 2007 to 2013, he was
a Specialist Research Scientist with the Finnish
Geospatial Research Institute. He is currently a
Professor at the School of Electronic Information
and Electrical Engineering, Shanghai Jiao Tong
University. He has authored or co-authored over
100 scientific papers. He is also an inventor of 25
patents and pending patents. His main research
is in the areas of indoor/outdoor seamless positioning, ubiquitous computing, wireless positioning, Bio-inspired navigation, context-aware applications, location-based services, and navigation of unmanned systems. Dr. Pei was a recipient of the Shanghai
Pujiang Talent in 2014 and ranked as the World’s Top 2\% scientists by
Stanford University in 2022.
\end{IEEEbiography}



\vfill

\end{document}